# A High-Level Model of Neocortical Feedback Based on an Event Window Segmentation Algorithm

Jerry R. Van Aken[*]

**ABSTRACT:** The author previously presented an *event window segmentation* (EWS) algorithm [5] that uses purely statistical methods to learn to recognize recurring patterns in an input stream of events. In the following discussion, the EWS algorithm is first extended to make predictions about future events. Next, this extended algorithm is used to construct a high-level, simplified model of a neocortical hierarchy. An event stream enters at the bottom of the hierarchy, and drives processing activity upward in the hierarchy. Successively higher regions in the hierarchy learn to recognize successively deeper levels of patterns in these events as they propagate from the bottom of the hierarchy. The lower levels in the hierarchy use the predictions from the levels above to strengthen their own predictions. A C++ source code listing of the model implementation and test program is included as an appendix.

In a previous paper [5], the author presented a segmentation algorithm that uses purely statistical methods to learn to recognize recurring patterns in an input stream of events. This algorithm will be referred to here as the *event window segmentation* (EWS) algorithm. An *event* is the arrival of an integer value (for example, a character code) in the input stream. For the purposes of this paper, the events in the input stream can be considered as ordered in time, but with possibly nonuniform intervals between successive events. A recurring pattern is an ordered set of adjacent events that occurs at a statistical frequency that is significantly higher than that expected for a random ordering of events.

The initial presentation of the EWS algorithm in [5] made no mention of the potential to use the statistical information acquired by the algorithm to make predictions about future events. The following discussion will focus on methods for extending the EWS algorithm to make such predictions.

This work was inspired by the description of the *memory-prediction framework* by Hawkins and Blakeslee [4], who theorize that the basis of human intelligence is the ability of each small region of the neocortex to learn to recognize recurring patterns. They propose that the primary function of the neocortex is to use the patterns stored in these regions to continuously make reliable predictions about future events. Furthermore, by connecting several such regions to form a hierarchy, successively higher regions in the hierarchy learn to recognize successively deeper levels of patterns in the event streams that enter the regions located at the bottom of the hierarchy. Each region in the hierarchy then uses the predictions from the regions above and below it to strengthen its own predictions.

The EWS algorithm is a convenient tool for experimenting with various types of feedback that might be exchanged between the levels in a hierarchy of regions, and for tuning this feedback to improve the predictive power at each level. This paper describes the operation of a high-level model in which each region in the hierarchy is implemented as an instance of the EWS algorithm. To support this model, the original algorithm is first adapted to predict future events in an input stream of events. These predictions are based on statistical information that is already gathered by the original algorithm. Next, the algorithm is extended to communicate with other regions in a hierarchy. The EWS algorithm in each region sends its local predictions to other regions, and uses feedback from these regions to strengthen its own predictions.

[*] Send correspondence to: Jerry Van Aken, Microsoft Corporation, One Microsoft Way, Redmond, WA 98052.

The hierarchy discussed by Hawkins and Blakeslee [4] is a tree-like hierarchy, as shown in Figure 1(a). For the sake of simplicity, the following discussion will describe a model hierarchy, as shown in Figure 1(b), in which each level in the hierarchy consists of a single region. Further extension of the EWS algorithm to support more complex hierarchies, such as the one in Figure 1(a), is a potential area for future work.

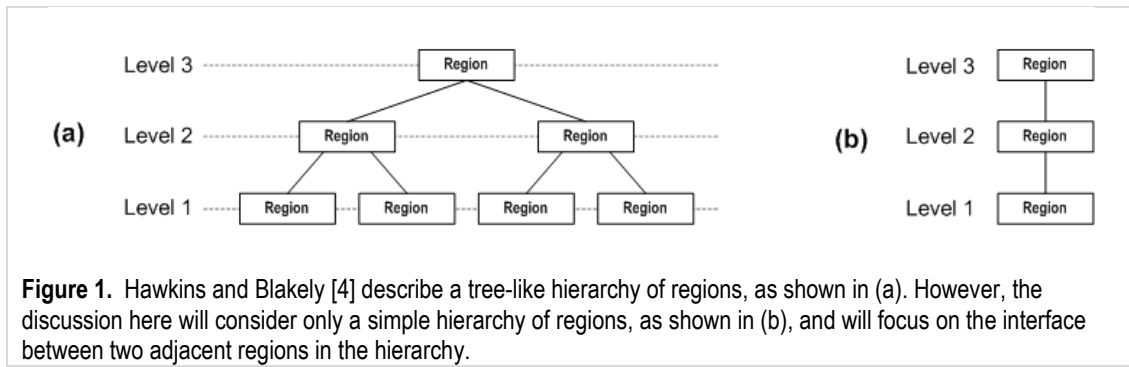

**Figure 1.** Hawkins and Blakely [4] describe a tree-like hierarchy of regions, as shown in (a). However, the discussion here will consider only a simple hierarchy of regions, as shown in (b), and will focus on the interface between two adjacent regions in the hierarchy.

All processing performed by the model hierarchy in Figure 1(b) is initiated by the arrival of an event in the input stream to level 1, at the bottom level of the hierarchy. This event always triggers processing activity in level 1, and this activity may or may not trigger activity in level 2, the next-higher level. If activity is triggered in the level 2, this activity may or may not trigger activity in level 3, and so on. If a prediction made at the top level is accurate, this prediction remains invariant for a relatively long period, during which processing activity is mostly confined to the lower levels in the hierarchy. If an unexpected event occurs at a lower level, activity quickly propagates to the top of the hierarchy.

The EWS algorithm does not try to simulate the operation of the biological neurons in the neocortex . However, if the basic operation performed by a human neocortical region is indeed to recognize recurring sequences, as proposed in [4], then the neocortex has at least this much in common with the EWS algorithm. Thus, understanding how to use feedback to improve the predictive power of a hierarchy of instances of the EWS algorithm might provide insights into the role of feedback in the neocortex.

## Overview of EWS algorithm

During the learning phase, the EWS algorithm receives an input stream of integer values. For example, these values can be character codes, and the input stream can consist of concatenated character strings, each of which is a word. This stream might be thousands of words in length. A portion of a stream that consists of words selected from the text of the Gettysburg Address—but in random order—might look like this:

      …requaltheircanearththataltogetherinscorefarlastinmenw…

Each word from the text appears many times in the stream, but the stream contains no markers to indicate the boundaries between words. The arrival of each character in the stream is referred to as an *event*. Each such event has an associated value, which, in this example, is a character code.

At the end of the learning phase, the algorithm has learned to segment the stream as follows:

...r_equal_their_can_earth_that_altogether_in_score_far_last_in_men_w...

That is, the algorithm can identify the boundaries between words and mark them (in this example, by inserting underscore characters).

*Sequence memory*
The EWS algorithm learns to recognize recurring sequences by building a data structure—called the *sequence memory*—that contains representations of the recurring sequences that are observed in the input stream. The algorithm avoids trying to store all possible sequences that might be constructed from the events in the stream because doing so would quickly overrun all available memory. Instead, the algorithm controls the growth of the sequence memory by storing a sequence *abcd* only after a previously stored sequence *abc* is frequently observed to be immediately followed by event *d*.

For example, if *abc* is a known recurring sequence that appears in the input stream 100 times, and the sequence *abcd* appears 20 times, this relatively high frequency indicates that the recurring sequence *abcd* is likely to be a word or part of a word. Of course, these same statistics can be used to predict future events: When an instance of *abc* is observed in the input stream, the probability that the next input event is *d* is P = 0.20.

If the sequence memory contains a sequence *abcd*, it also contains all subsequences of *abcd*. That is, it contains *abc*, *bcd*, *ab*, *bc*, *cd*, *a*, *b*, *c*, and *d*.

*Event window*
The EWS algorithm uses a second data structure, called the *event window*, that stores a small number (typically, less than 16) of the most recent events received from the event stream. The algorithm tries to find a segmentation of the events in the event window that simultaneously identifies all of the word boundaries in these events. This technique is more powerful than trying to identify an individual word in isolation without taking into account the words that surround it [5].

As each new event arrives in the input stream, the algorithm appends this event to the right edge of the event window, as shown in Figure 2(a). Then the algorithm uses the statistical information in the sequence memory to calculate the most probable word segmentation of all the events (including the newest event) in the event window.  If the algorithm finds a highly probable segmentation, in which case the leftmost word in the segmentation can be reliably identified, this word is detached from the event window, as shown in Figure 2(b). (The algorithm can sometimes detach more than one word in response to a new event.) Otherwise, the algorithm does nothing further until the next input event arrives.

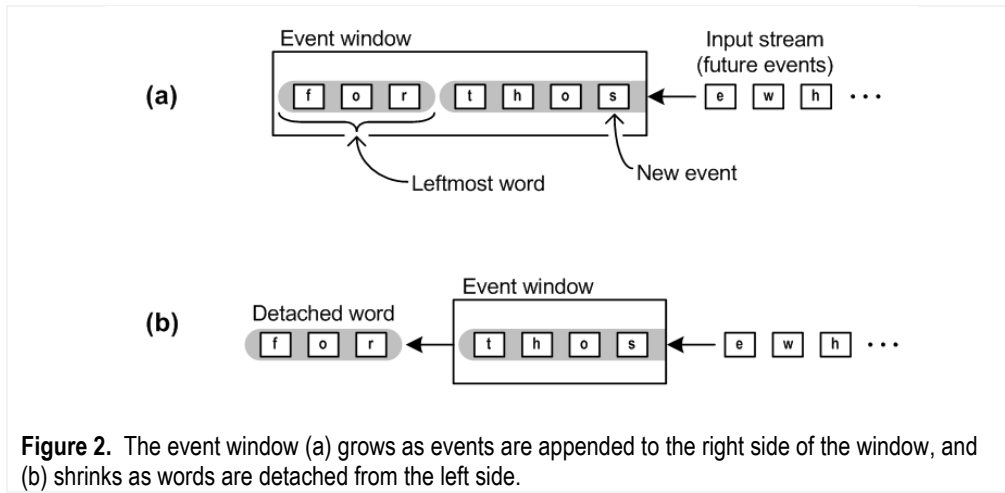

**Figure 2.** The event window (a) grows as events are appended to the right side of the window, and (b) shrinks as words are detached from the left side.

Over time, the event window alternately grows and shrinks as events are added to the right edge and words are detached from the left edge. If the algorithm successfully identifies the words to detach, the left edge of the event window is always aligned to a word boundary.

By segmenting an event window that might contain multiple words, the EWS algorithm can resolve ambiguities that might otherwise cause segmentation errors. For example, the Gettysburg Address contains both of the following phrases:

- "for those who here gave their lives"
- "forth on this continent"

Thus, if the input stream consists of the words in the Gettysburg Address, in their proper order but without boundary markers between words, the event window might, at some point in the stream, contain the following six events:

  fortho...

The algorithm is designed to keep the left edge of the event window always aligned to a word boundary. Thus, the event (letter 'f') at the left edge is highly likely to be the start of a word. This word must be either "for" or "forth" if the choice of words is restricted to those that occur in the Gettysburg Address. In the absence of any additional information, however, the algorithm cannot determine which of these two words to detach, so the algorithm does nothing and waits for the next input event from the input stream. If the next event is the letter 's', the event window will contain the following seven events:

  forthos...

At this point, the only possible segmentation of the event windows contents is "for_thos...", where the underscore character marks the word boundary. The segmentation "forth_os..." must be rejected because no word in the Gettysburg Address starts with "os". The algorithm now confidently identifies the leftmost word in the event window as "for" and detaches this word from the event window. After detaching this word, the event window contains just these four events:

> thos...

This process continues as more events are appended to the right side of the event window, and more words are detached from the left side.

## Building a hierarchy

As previously discussed, several instances of the EWS algorithm can be connected to form a hierarchy. Figure 3 shows two adjacent levels in such a hierarchy. At the bottom of the figure, an input stream of events enters level 1, which is the lower of the two levels.

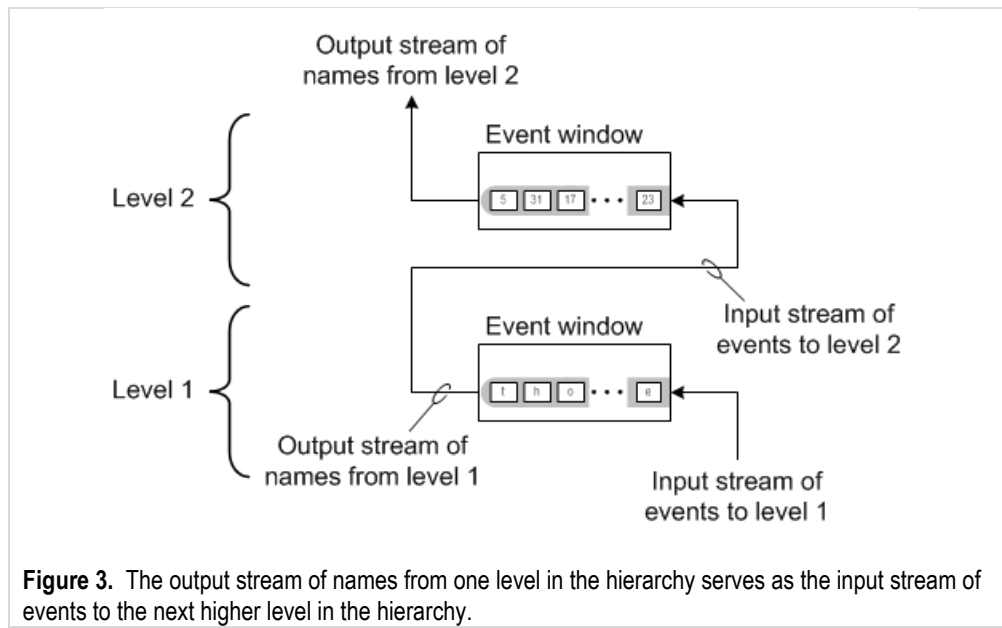

**Figure 3.** The output stream of names from one level in the hierarchy serves as the input stream of events to the next higher level in the hierarchy.

The succession of words detached from the event window in level 1 is transformed to an input stream of events that is sent to the next higher level in the hierarchy. Each word that is detached from the event window in level 1 is assigned a *name*, which is simply an integer value. For example, the event window in Figure 2(b) might assign the name 23 to the word "for". These names serve as the events in the input stream to the next higher level.

The assignment of names to words is not part of the original EWS algorithm [5], but is a necessary addition to support hierarchical learning, as described in [4].

In Figure 3, the only buffering of events occurs in the event windows. There is no buffering of events between windows. Thus, when a word is detached from the event window in level 1, the name of this word propagates without delay to the event window of the next higher level.

Learning starts at the bottom of the hierarchy and proceeds upward, one level at a time. During the learning phase at each level, name assignment at this level is turned off, which means that the next-higher level receives no input stream. Only after a particular level finishes its learning phase is name assignment at this level turned on so that the next-higher level can start its learning phase.

After name assignment is turned on at a particular level in the hierarchy, the EWS algorithm at this level checks each word that is detached from the event window to see if this word has already been assigned a name. If not, the algorithm assigns a unique and permanent name to the word.

**Incorporating feedback**

After two successive levels in the hierarchy have both completed their learning phases, they can start to exchange feedback. In Figure 3, for example, level 1, the lower of the two levels, might first learn to recognize the *words* in the Gettysburg Address, and then teach level 2 to recognize *phrases* (sequences of words) in the Gettysburg Address. After both levels have completed their learning, they can make useful predictions about future events in their respective input streams, as described previously. Level 1 can predict the next character in its input stream, and level 2 can predict the next word in its input stream.

The accuracy of these predictions tends to vary from one event to the next. In Figure 3, level 1 does poorly at predicting the first one or two letters at the start of a new word but is much better at predicting the letters near the ends of the words it has learned. Similarly, level 2 does poorly at predicting the first one or two words at the start of a new phrase but is much better at predicting words near the ends of the phrases it has learned. This phenomenon is described by Elman [2][3].

After level 1 receives the last character in a word, this level is typically unable to accurately predict the character at the start of the next word in its input stream. However, level 2 might be able to confidently predict the next event (a word) in *its* input stream and send this prediction down to level 1. Of course, level 2 knows only the *names* of words, so when it sends the name of the predicted word to level 1, level 1 must translate this name back into the sequence of characters in the word. After performing this translation, level 1 can confidently predict the next character in its input stream.

Similarly, information provided by the lower of two adjacent levels in the hierarchy can improve the predictions made by the level above. In the preceding example, if level 2 just received (the name of) the word "that" in its input stream, but has no other contextual information for making a prediction, the EWS algorithm that runs in level 2 might have difficulty choosing among several possible local predictions for the next word in the stream. The reason is that the Gettysburg Address contains 13 instances of the word "that", and each instance is followed by a word from the following set:

> { all, cause, field, from, government, nation, that, these, this, war }

Level 2 needs additional information to reliably predict which of these words will be next.

Fortunately, after level 1 receives from its input stream the first character in the next word, the number of possibilities is significantly reduced. For example, if the first character is 'f', level 1 can narrow down the possibilities to the following words:

> { far, fathers, field, final, fitting, for, forget, forth, fought, four, freedom, from, full }

This is the set of words in the Gettysburg Address that start with 'f'. Level 1 transmits this set of possible next words to level 2, which determines the intersection of this set with its local set of predicted words. The following two words are the only ones that appear in both sets:

> { field, from }

Thus, by using the feedback from level 1, level 2 has narrowed the number of possible next words from ten to two. When level 1 receives the second character in the next word, and the character is 'i', level 1 can further narrow the list of possibilities to words that start with "fi", and level 2 can use this information to confidently predict that the next word is "field".

**Representing sequences**

Figure 4 shows a part of a sequence memory that contains the recurring sequences of letters in the Gettysburg Address. The part shown in the figure is the branch of the sequence memory that contains all the sequences (including both words and parts of words) that start with the letter 'b'. The following sequences are stored in this branch:

> "berty", "before", "brought", "brave", "birth", "bly", "bove", "battlefield", "by"

Some of these sequences—like "before" and "brought"—are complete words. Others—like "berty" and "bly"—are parts of words.

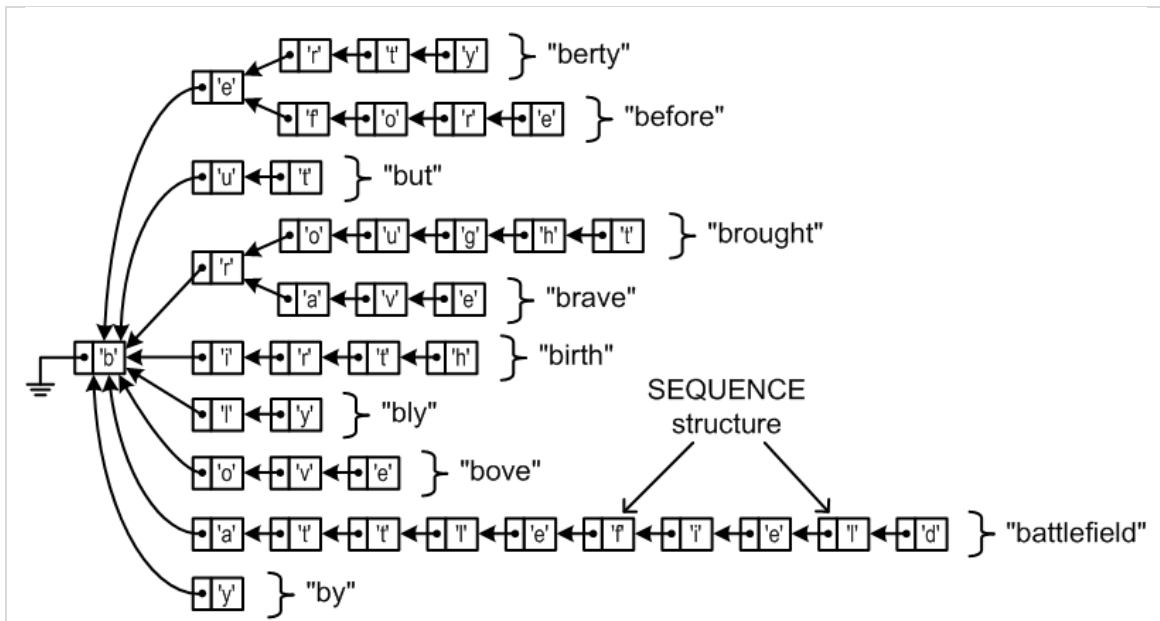

**Figure 4.** The sequence memory is made up of linked SEQUENCE structures. This example shows the branch of sequence memory that contains all recurring sequences that start with the letter 'b'.

Each small rectangle in Figure 4 is a SEQUENCE structure that contains both a link (a pointer value shown as an arrow) to another SEQUENCE structure, and an event value (a character code in the range 'a' to 'z'). A SEQUENCE structure contains other information (such as statistical data) that, for simplicity, is not shown in the figure.

For example, if the sequence memory already contains the sequence "bu", the sequence "but" is formed by adding a SEQUENCE structure that contains the letter 't' and that points to the sequence structure that represents the sequence "bu".

To assign a name to the sequence "but", the algorithm uses a lookup table to associate the name (an integer value, as previously discussed) with a link to the SEQUENCE structure at the end of the stored sequence "but" (the one that contains the letter 't' in Figure 4). The characters in the sequence can be obtained (in reverse order) by traversing the links from one SEQUENCE structure to the next. In this way, level 1 in the hierarchy can convert (the name of) a word predicted by level 2 into a predicted next character in the input stream to level 1.

**Implementing the event window**

The event window is a two-dimensional data structure that is capable of representing all possible segmentations of an ordered sequence of events. For example, the event window shown in Figure 5 contains the events *a*, *b*, *c*, and *d*, where *a* is the oldest event in the window and *d* is the most recent. As shown in the following list, there are eight possible ways to segment these four events:

$$abcd, abc\_d, ab\_cd, ab\_c\_d, a\_bcd, a\_bc\_d, a\_b\_cd, a\_b\_c\_d$$

(As before, segment boundaries are marked by underscore characters.) These segmentations use various combinations of the following ten sequences: *abcd*, *abc*, *bcd*, *ab*, *bc*, *cd*, *a*, *b*, *c*, and *d*. These ten sequences are represented by the ten small black circles in Figure 5. For example, the sequence *abcd* is represented by the black circle in the top right corner of the figure, and the sequence *a* is represented by the circle in the bottom left corner.

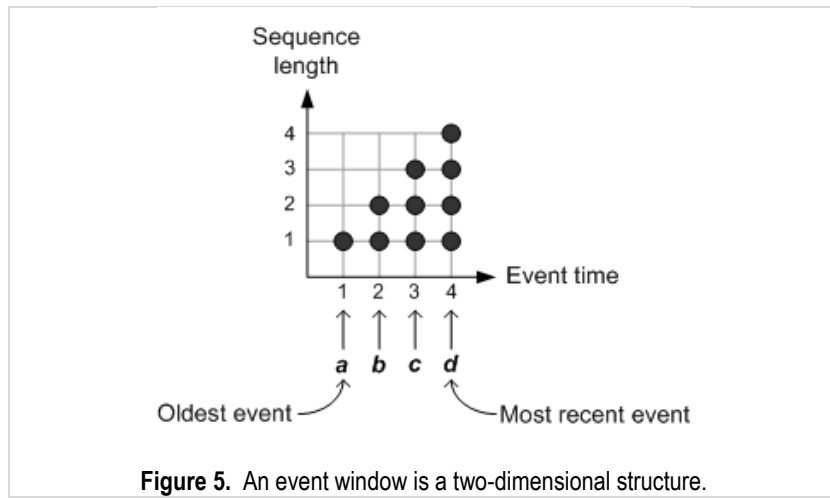

**Figure 5.** An event window is a two-dimensional structure.

In Figure 5, the horizontal dimension is event time, and the vertical dimension is sequence length. Each grid point in the structure is identified by its (*time*, *length*) coordinates. Table 1 lists the coordinates of each sequence represented by a black circle in Figure 5.

Table 1. The (*time*, *length*) coordinates for the sequences in Figure 5.

| (*time, length*) | Sequence |
|---|---|
| (4,4) | abcd |
| (3,3) | abc |
| (4,3) | bcd |
| (2,2) | ab |
| (3,2) | bc |
| (4,2) | cd |
| (1,1) | a |
| (2,1) | b |
| (3,1) | c |
| (4,1) | d |

For example, if the input stream consists of the words in the Gettysburg Address, in their proper order but without boundary markers between words, the event window might, at some point in the stream, contain the following eleven character codes:

>butinalarge...

Figure 6 shows the event window for this example.

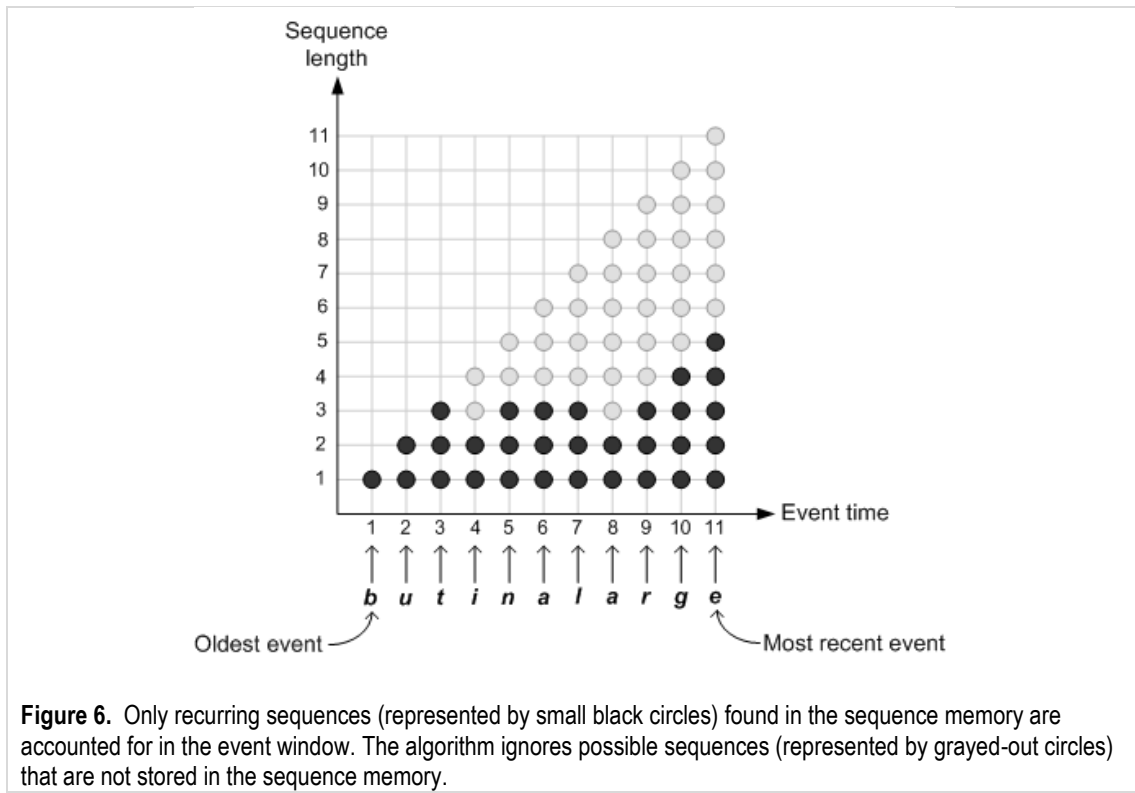

**Figure 6.** Only recurring sequences (represented by small black circles) found in the sequence memory are accounted for in the event window. The algorithm ignores possible sequences (represented by grayed-out circles) that are not stored in the sequence memory.

When the algorithm evaluates possible segmentations of the event window contents, it considers only combinations of sequences that are stored in the sequence memory. The small grayed-out circles in the upper half of Figure 6 indicate that the corresponding sequences are not found in the sequence memory and can therefore be ignored by the EWS algorithm. The black circles in the lower half of the figure represent sequences that are stored in the sequence memory.

In Figure 6, an event window cell that is marked with a black circle does not, in fact, contain a copy of the sequence information from the sequence memory. Instead, the cell contains a *pointer* to the node in the sequence memory that contains this information. If, at some time, an event window contains two or more instances of the same sequence, these instances all point to the same node in sequence memory.

The most probable segmentation of the event window contents in Figure 6 is shown in Figure 7. The shaded areas in Figure 7 surround the event window cells that are part of this segmentation. The pointers from these cells to the corresponding nodes in the sequence memory are shown as dashed arrows.

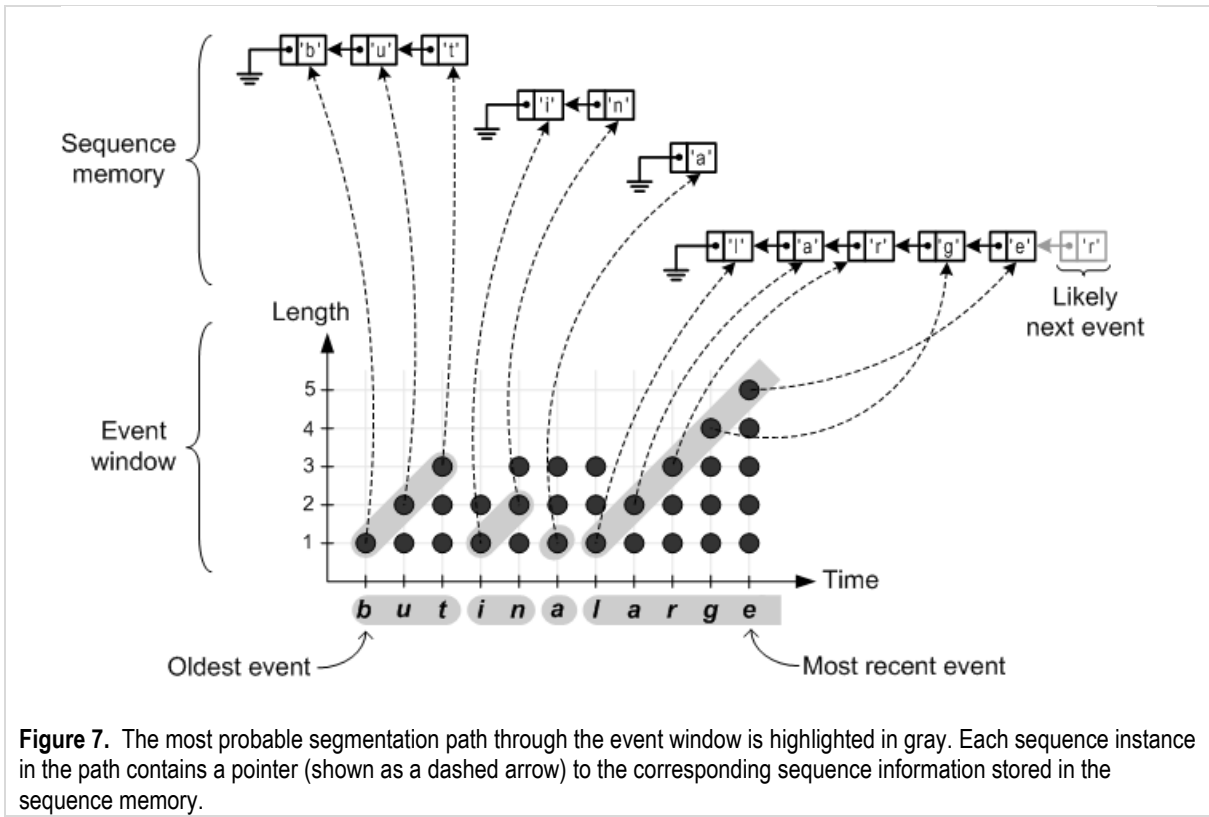

**Figure 7.** The most probable segmentation path through the event window is highlighted in gray. Each sequence instance in the path contains a pointer (shown as a dashed arrow) to the corresponding sequence information stored in the sequence memory.

For example, the event window cell that corresponds to the sequence "but" contains a pointer to a SEQUENCE structure that contains the character code 't' and that points to a list of SEQUENCE structures that represents the sequence "bu".

For a description of how to calculate the most likely segmentation of the event window contents, see [5].

## Making local predictions

An instance of the EWS algorithm that runs in a lower level in the hierarchy might be able to predict the next event in its input stream, but must rely on the levels above it to make predictions that are farther into the future. In the Gettysburg Address example described previously, if level 1 learns to predict the next *letter* in its input stream, then level 2 learns to predict next *word* in its stream. If a third level were to be added to this hierarchy, this level would learn to predict the next *phrase*, and so on. Thus, successively higher levels in the hierarchy can make useful predictions about successively more distant future events, and then feed these predictions downward to the lower levels.

However, predictions from above might unavailable because either there is no level above, or the level above has not yet learned to make predictions. There might also be times when the level above does makes a prediction but then assigns this prediction a low level of confidence.

If a particular level does not receive a viable prediction from above, this level must make its own *local prediction*. That is, the EWS algorithm at this level is forced to rely only on the contents of its event window—and, possibly, feedback from the level below—to make predictions about future events in its input stream.

When making a local prediction, the extended EWS algorithm tries to predict only the very next event in its input stream. To avoid complexity, the algorithm does not try to predict more than one event into the future.

For the purpose of making local predictions, the extended EWS algorithm attaches a *prediction column* to the right edge of the event window, as shown in Figure 8. The event window in this example contains the same set of events as the examples in Figures 6 and 7. In Figure 8, the black circles in the prediction column are predicted sequences of various lengths. For example, the sequence of length 6 in the prediction column is the predicted successor to the sequence of length 5 ("large") in the rightmost column of the event window.

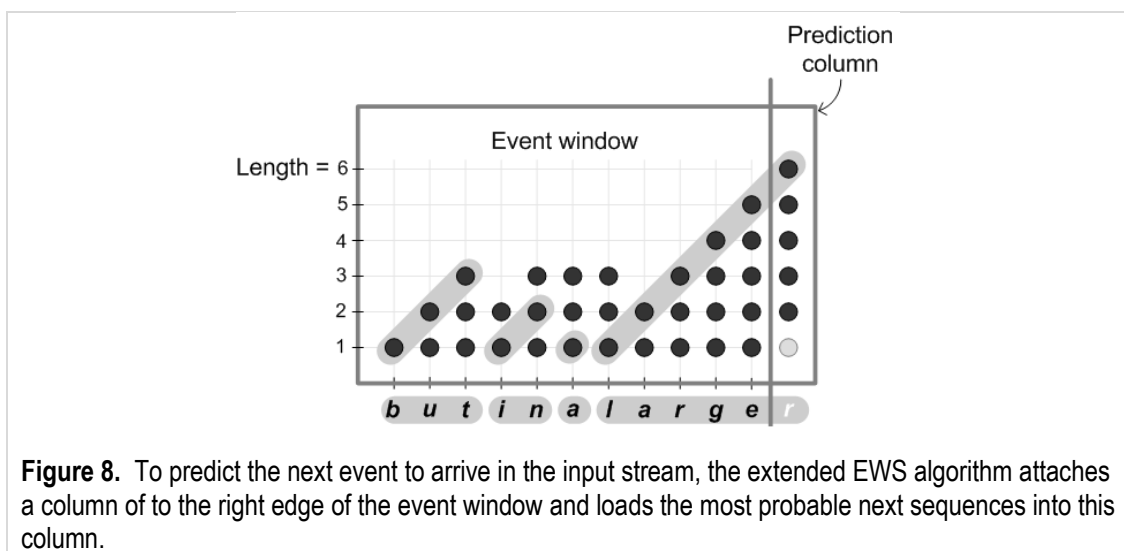

**Figure 8.** To predict the next event to arrive in the input stream, the extended EWS algorithm attaches a column of to the right edge of the event window and loads the most probable next sequences into this column.

To use the prediction column to predict the next event in the input stream, the EWS algorithm follows these steps:

1. Identify the most probable next sequence of each length (with the possible exception of length 1, as will be explained) and load this sequence into the corresponding position in the prediction column. If L is the length of the longest sequence in the rightmost column of the event window, the length of the longest possible sequence in the prediction column is L+1.
2. For each sequence in the prediction column, calculate the most probable segmentation of the event window contents that ends in this sequence. The probability of this segmentation is recorded so that it can be used in step 3.
3. Compare the probabilities recorded in step 2, and select the sequence in the prediction column that has the most probable segmentation.

The sequence selected in step 3 determines what the predicted next event in the input stream will be. The predicted next event is simply the last event in this sequence.

For example, in Figure 8, the most recent event to arrive in the input stream is the letter 'e', which is shown at the bottom of the rightmost column in the event window. The sequences (shown as black circles) in this column are as follows (in top-down order):

"large", "arge", "rge", "ge", and "e"

The EWS algorithm that runs at this level populates the prediction column with the sequences that are the most likely successors to the preceding sequences. In Figure 8, the prediction column contains the most likely successor sequences of lengths 2, 3, 4, 5, and 6. For example, the most likely successor to "large" is "larger", which the algorithm selects as the sequence of length 6 in the prediction column shown in the figure. (If the level below has sent information that eliminates certain sequences from further consideration, the algorithm ignores these sequences during the selection of the most likely successors.)

Next, for each successor sequence in the prediction column, the algorithm determines the most probable segmentation of the event window contents that ends in this sequence. (The "first-level scoring" method described in [5] is used to find the most probable segmentation.) In the example in Figure 8, the following segmentation is the most probable:

but_in_a_larger

The last character, 'r', in this segmentation is added by the winning sequence in the prediction column. That is, the local prediction for the next character in the input stream is 'r'. The algorithm records the probability associated with this prediction and sends this probability downward with the prediction to indicate how reliable the prediction is.

The EWS algorithm must account for the possibility that the right edge of the event window is aligned to a word boundary. If the next event to arrive in the input stream will be the start of a new word, the algorithm is unable to make a useful local prediction. Therefore, unless a viable prediction is available from above, the algorithm sends no prediction (or, equivalently, sends a null prediction) downward in the hierarchy. In Figure 8, the sequence at the bottom of the prediction column is grayed out to indicate that the algorithm declines to try to locally predict the next event if this event is the start of a new word.

In a hierarchy, the EWS algorithm that runs in a lower level is likely to receive useful predictions from the level above. A strong prediction from above overrides the lower level's local prediction. The prediction from above is especially helpful if the next event in the local input stream is the start of a new word.

**Predictions based on soft events**

In a hierarchy of EWS algorithm instances, the event window at each level of the hierarchy can be outfitted with a prediction column (see Figure 8) so that the algorithm at this level can make predictions about future events in its local input stream.

In Figure 3, the output stream from level 1 in the hierarchy forms the input stream to level 2. If, as in previous examples, the input stream to this level consists of the letters in the Gettysburg Address, this level can learn to predict the next letter in its input stream. Similarly, level 2 can learn to predict the next word (or, more precisely, the *name* of the next word) in its input stream.

However, Figure 3 implies that the next word in the input stream to level 2 is always the next word to be *detached* from the event window in level 1. Thus, level 2 predicts the arrival of a word that level 1 typically has already received, and there is little value in predicting events that have occurred in the past. What is needed is a way for level 2 to predict the name of the next word to arrive in the input stream to level 1 *before* it arrives. To do so, level 2 must look ahead by peeking at the words *inside* the event window in level 1 before these words are detached from the window.

In the example shown in Figure 9, the event window in level 2 contains seven events. The five oldest events (starting from the left side of the window) are the names of words that have been detached from the event window in level 1. However, the two most recent events (on the right side of the event window in level 2) are the names of words that have not yet been detached from the event window in level 1. The algorithm in level 1 has *tentatively* identified these words based on its current segmentation of the events in its event window. This segmentation is a best guess based on current information and might change when the next event arrives in the input stream to level 1.

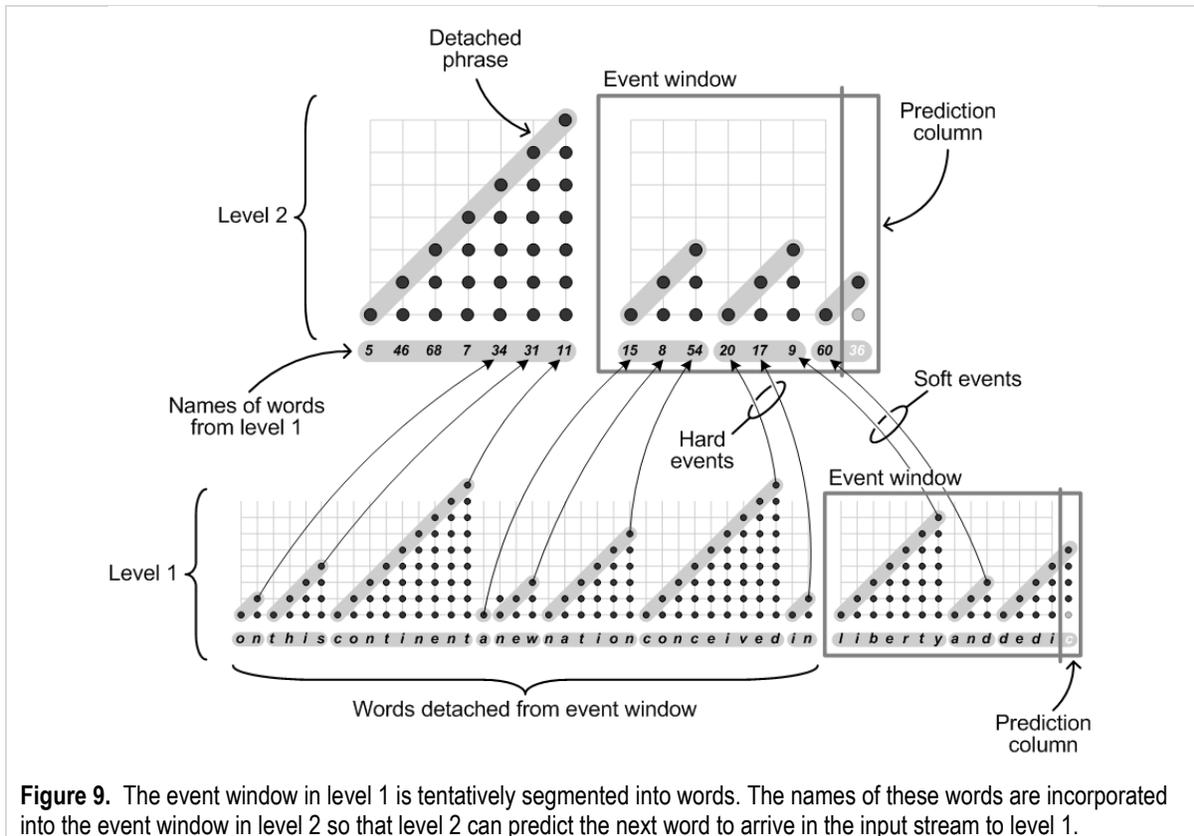

**Figure 9.** The event window in level 1 is tentatively segmented into words. The names of these words are incorporated into the event window in level 2 so that level 2 can predict the next word to arrive in the input stream to level 1.

These tentative events are referred to as *soft events* to distinguish them from the *hard events* that are based on words that have already been detached from the event window in level 1. The decision to detach a word is final, meaning that the algorithm never considers changing the segmentations of words that have already been detached. Although the soft events from level 1 *might* change, they typically are reliable enough that level 2 can use them to make solid predictions.

**Filtering feedback**
In Figure 9, levels 1 and 2 use prediction columns to predict the next events in their respective input streams. As previously discussed, level 1 will ignore the local prediction from its prediction column if a strong prediction is available from level 2.

In the upper right corner of Figure 9, the prediction column in level 2 tries to predict the next event in the input stream to level 2, which is the word that follows the two soft events that represent the words "liberty" and "and". To corroborate this local prediction, level 2 will use information from level 1, if such information is available, to eliminate from further consideration any predictions that are inconsistent with the most recently received events in the event window in level 1.

As shown in the lower right corner of Figure 9, the four most recent events in the event window in level 1 are the letters "dedi". These events follow the tentatively identified words "liberty" and "and" in the window. Level 1 can use these four events to help level 2 narrow down the list of possible next words that might follow "liberty" and "and".

Continuing with the example in Figure 9, level 1 must consider the possibility that the next word is "d", "de", "ded", or "dedi". However, none of these sequences is a word in the Gettysburg Address, so—for this example—level 1 can be immediately rule them out as possible next words.

The remaining possibility is that the next word is five or more letters in length and starts with the letters "dedi". Indeed, two words—"dedicate" and "dedicated"—in the Gettysburg Address fit this description. Therefore, level 1 determines that the next word will be one of these two words, and sends this information to level 2 in the form of a *feedback filter*. A feedback filter specifies all possible next words that are consistent with the events received thus far. For this example, the filter specifies the words "dedicate" and "dedicated".

The EWS algorithm implements a feedback filter as a bitmask, which is a simple, one-dimensional array of bits. Each bit in the filter represents a name that the algorithm has assigned to a word. For example, if level 1 in Figure 9 has assigned the names (integer values) 48 and 36 to the words "dedicate" and "dedicated", bits 48 and 36 in the filter are set to one to indicate that the name of the next word will be one of these two values. All other bits in the filter are set to zero. If level 1 has assigned names 1 to N to the words in its output stream, the filter is implemented as a bitmask that is at least N bits in length.

The feedback filter from level 1 is used by level 2 to eliminate from further consideration any of the level-2 local predictions that are not specified in the filter. This elimination occurs before level 2 uses its prediction column to predict the most likely next event in its input stream. Thus, the prediction that level 2 sends to level 1 will be consistent with the feedback filter that level 1 previously sent to level 2.

The EWS algorithm in level 1 issues a soft event only for a word that has been received in its entirety into the event window. In the preceding example, level 2 will choose to predict that either "dedicate" or "dedicated" will be the next word. Only after the last letter of the predicted word is received into the level-1 event window is the soft event representing this word sent to level 2.

Thus, if level 2 is predicting that the next word is "dedicate", the soft event representing this word is sent when the letter 'e' at the end of "dedicate" is appended to the event window. However, if level 2 is predicting that the next word is "dedicated", the soft event representing this word is sent when the letter 'd' at the end of this word is appended to the event window. As long as level 2 succeeds in making correct predictions, it is always in a position to predict the next word before the first letter in this word arrives in the input stream to level 1.

The list of words in a feedback filter frequently includes words that have already been received in their entirety into the event window. For example, if the event window in level 1 contains the letters "thei" and level 2 is strongly predicting that the next word will be "their", the feedback filter must still account for the possibility that the event window contains the word "the" followed by the first letter in a word that starts with the letter 'i'. In fact, several words in the Gettysburg Address start with 'i', so this possibility cannot be dismissed. Thus, the words in the feedback filter must include both "the" and "their". However, if the next letter to arrive in the input stream is 'r', the word "the" can, at this point, be omitted from the feedback filter. The reason is that "the" is a viable possibility only if the characters that follow this word in the event window are *segmentable*—that is, they can be segmented into words in the Gettysburg Address lexicon. The two letters "ir" are segmentable if either (1) a word in the lexicon starts with the letters "ir", or (2) "i" is a single-letter word in the lexicon and is followed by a word that starts with the letter 'r'. However, neither of these options is viable, and so the only word that level 1 includes in the new feedback filter sent to level 2 is "their". On receipt of this new filter, level 2 confidently predicts that the next word is "their". Level 1 responds to this prediction by sending a soft event representing the word "their" to level 2 so that level 2 can predict the word that follows "their".

**Recovering from prediction errors**

Level 2 is not always able to make predictions with a high level of certainty. A prediction made with an 80 percent probability of being right has a 20 percent chance of being wrong. When a prediction is wrong, the hierarchy must recover quickly so that it can resume making reliable predictions with as little disruption as possible.

As previously discussed, level 2 is most likely to make prediction errors in the first word or two of a new phrase. Consider the example of a test program that randomly selects the phrase "are created equal" from the Gettysburg Address to present as an input to level 1. That is, the input stream to level 1 will be the sequence "arecreatedequal". Because this phrase is chosen at random, level 2 has no context to use to predict the first word in this sequence until level 1 has received the first letter in the sequence. However, if recent predictions from level 2 have thus far been accurate, level 1 will at least know that the first letter in this sequence, when it does arrive in the input stream, is highly likely to be the start of a new word.

After the first letter, 'a', in the example sequence "arecreatedequal" arrives in the input stream to level 1, this level adds this letter to its event window, and then builds a feedback filter that includes all words in the Gettysburg Address that start with the letter 'a'. Level 1 sends this filter to level 2, which responds by selecting the most likely word from this filter to be the prediction sent downward to level 1. Of the 24 word instances in the Gettysburg Address that start with the letter 'a', the single-letter word "a", which appears seven times, is the most likely. Thus, level 2 predicts that the next word is "a", and assigns a probability of about 7/24 to this prediction. In response to this prediction, level 1 sends a soft event representing the word "a" to level 2, and level 2 then tries to predict the word that will follow the presumed word "a". At this point, levels 1 and 2 are both proceeding on the incorrect assumption that the first word in the sequence is "a" instead of "are".

The next letter to arrive in the input stream to level 1 after 'a' is 'r'. Based on the prediction from level 2, level 1 assumes that 'r' is the first letter in a new word and sends level 2 a feedback filter that contains all the words in the Gettysburg Address that start with 'r'. In response, level 2 selects the most probable word in the filter to be the predicted word that follows the incorrectly identified word "a". It is not yet obvious to levels 1 and 2 that they are being led down the "garden path" to a dead end.

The error finally becomes obvious after the third letter in the sequence, 'e' arrives in the input stream to level 1. At first, level 1 is still unaware of the error and sends level 2 a feedback filter containing all the words from the Gettysburg Address that start with the letters "re". However, no instance of a word that starts with "re" in the Gettysburg Address is preceded by the word "a". Thus, level 2 is unable to make a prediction for the next word—or, equivalently, level 2 sends a null prediction with a confidence level of zero to level 1.

This low confidence level informs level 1 that the tentative segmentation of the first three letters, "are", in the sequence is wrong, and level 1 must now initiate a recovery from this error. In the current implementation of the EWS algorithm, level 1 recovers by *retracting* the most recent soft event. This retraction has the following effects:

- The soft event that incorrectly indicates that the first word in the sequence is "a" is simply removed from the event window in level 2.
- All three letters, "are", in the event window in level 1 are used to construct a new feedback filter.

The new feedback filter constructed by level 1 initially contains the words "a" and "are". However, level 1 remembers that the word "a" has been retracted and deletes this word from the filter before the filter is sent to level 2. Level 2 then correctly predicts that the first word in the sequence is "are".

**Training and testing the model**

In the preceding discussion, each level in a hierarchical, high-level model of the neocortex is implemented as an instance of the EWS algorithm. This discussion has focused on the communication between two adjacent levels in the hierarchy. As previously explained, the levels in the hierarchy are trained starting from the bottom up, one level at a time.

If level 1 is the bottom level in the hierarchy, the training of the next-higher level, level 2, starts only after level 1 has completed its training. Initially, learning is enabled only in level 1. During the learning phase in level 1, the input stream to this level is generated by a test program. The events in this stream consist of character codes. To generate this stream, the test program randomly selects words from a hidden dictionary and appends the characters in each word to the stream. The following C++ program generates such a stream from the words in the Gettysburg Address:

```
#include <stdlib.h>
#include <stdio.h>

char *test1[] =
{
    "four", "score", "and", "seven", "years", "ago", "our", "fathers",
    "brought", "forth", "on", "this", "continent", "a", "new",
    "nation", "conceived", "in", "liberty", "and", "dedicated", "to",
    "the", "proposition", "that", "all", "men", "are", "created",

    ...

    "shall", "not", "perish", "from", "the", "earth"
};

const int NUM_TEST_WORDS = sizeof(test1)/sizeof(test1[0]);
const int NUM_LEARNING_WORDS = 150000;

void main()
{
    for (int ix = 0; ix < NUM_LEARNING_WORDS; ++ix)
```

```
    {
        int jx = rand() % NUM_TEST_WORDS;
        printf("%s", test1[jx]);
    }
}
```

The following characters are a small part of the event stream generated by this program:

```
thattestingthisdevotiontheourproperlivefarasnewthatshallallthatgavenewlibertygaveco
nceivedthethatcanheretonoblyrestinginfromcanaresopowercannationhereuscontinentthese
toadeadneverfreedomaonnotmenaaltogetherfieldandthehavethatbravecanofforgetnotthewei
ncauseinallthatearthpropositionitawededicatedofoftothisthesesevenallnotnevernotbeon
noblynationgreatdedicatetotheisnotincreasednotoherefarforondevotiontheearethistak
etocantohaveisworldandandbutciviltotheydeadtheyfreedomtothiswhichhonoredfarsensepla
cewhohavestruggledbirthcomeaddanylittlewhichdevotiondedicatethedevotiononisweliversw
hetherhaveresolveandbuttheyitnewourthereberebethatofwarequalcomedeadandincreasedadddbi
rthmeasurehaveforresolvepeopledevotiontheywhothehereaiscanfittingwewhoendurebutcons
ecratedallandshallthattherethatbypeople...
```

After level 1 finishes its training, but before the training of level 2 starts, level 1 can do the following:

- Identify the boundaries between successive words in the input stream.
- Make local predictions about what the next character in the input stream will be.

After training, the test program might send the following stream to level 1:

```
fourscoreandsevenyearsagoourfathersbroughtforthonthiscontinentanewnationconce
ivedinlibertyanddedicatedtothepropositionthatallmenarecreatedequalnowweareeng
agedinagreatcivilwartestingwhetherthatnationoranynationsoconceivedandsodedica
tedcanlongenduredwearemetonagreatbattlefieldofthatwarwehavecometodedicateaport
ionofthatfieldasafinalrestingplaceforthosewhoheregavetheirlivesthatthatnation
mightliveitisaltogetherfittingandproperthatweshoulddothisbutinalargersensewec
annotdedicatewecannotconsecratewecannothallowthisgroundthebravemenlivingandde
adwhostruggledherehaveconsecrateditfaraboveourpoorpowertoaddordetracttheworld
willlittlenotenorlongrememberwhatwesayherebutitcanneverforgetwhattheydidherei
tisforustherelivingrathertobededicatedheretotheunfinishedworkwhichtheywhofought
herehavethusfarsonoblyadvanceditisratherforustobeherededicatedtothegreattaskr
emainingbeforeusthatfromthesehonoreddeadwetakeincreaseddevotiontothatcausefor
whichtheygavethelastfullmeasureofdevotionthatweherehighlyresolvethatthesedead
shallnothavediedinvain...
```

Level 1 can convert this input stream into an output stream in which the word boundaries in the stream are marked by spaces, as follows:

```
four score and seven years ago our fathers brought forth on this continent a
new nation conceived in liberty and dedicated to the proposition that all men
are created equal now we are engaged in a great civil war testing whether
that nation or any nation so conceived and so dedicated can long endure we
are met on a great battlefield of that war we have come to dedicate a portion
of that field as a final resting place for those who here gave their lives
that that nation might live it is altogether fitting and proper that we
should do this but in a larger sense we can not dedicate we can not
consecrate we can not hallow this ground the brave men living and dead who
struggled here have consecrated it far above our poor power to add or detract
the world will little note nor long remember what we say here but it can
never forget what they did here it is for us the living rather to be
dedicated here to the unfinished work which they who fought here have thus
far so nobly advanced it is rather for us to be here dedicated to the great
task remaining before us that from these honored dead we take increased
```

```
    devotion to that cause for which they gave the last full measure of devotion
    that we here highly resolve that these dead shall not have died in vain...
```

Level 1 can also try to predict the next character to arrive in the stream. Before each character is fed to level 1, the test program can compare this character with the character predicted by level 1. In the following output stream, the test program highlights each *incorrectly* predicted character by converting the character from lower case to upper case:

```
FoUrSCoreANdSeVenYeArsAGoOUrFAThersBRoughtForthOnthIsCOnTinentANEWNAtionCOnCe
ivedInLIBertyANdDedicatedTOThePRopositionThAtALlMenAReCReaTedeQualNoWWeAReeNG
agedInAGReatCIvilWArTEStingWHEtherThAtNAtionORANYNAtionSOCOnCeivedANdSODedica
tedCaNLOngeNDureWeAReMeTOnAGReatBAttlefieldOFThAtWARWeHAVeCOMeTODedicateAPORt
ionOFThAtFIEldASAFInAlReSTingPLAceForthOSeWHoHereGAveTheIrLIvEsThAtthAtNAtion
MIghtLIvEITISALTogetherFITtingANdPRopErThAtWeSHOuldDOThIsBUtInAlARGerSeNseWeC
aNNotDedicateWeCaNNotCOnsecrateWeCaNNotHALlowThIsGROundTheBRAveMenLIvingANdDe
AdWHoSTRuggledHereHAVeCOnsecratedITFArABoveOUrPOOrPOWerTOADdORDeTractTheWORld
WIllLITtleNoteNORLOngReMemberWHAtWeSAyHereBUtITCaNNEVerForGetWHAtTheYDIDHereI
TISForUSTheLIvingRAtherTOBeDedicatedHereTOTheUNFinishedWOrKWHIchTheYWHoFoUGht
HereHAVeThUsFArSONoBlyADVancedITISRAtherForUSTOBeHereDedicatedTOTheGreatTAskR
eMAiningBeForeUSThAtFROmTheSeHOnoredDeAdWeTAKeInCreasedDeVotionTOThAtCaUseFor
WHIchTheYGAveTheLAStFUllMeAsureOFDeVotionThAtWeHereHIGhlyReSolveThAttheSeDeAd
SHallNotHAVeDIEdInVAinT...
```

Level 1 does a poor job of predicting the first letter or two at the start of a word, but is much better at predicting the letters toward the ends of words. As previously mentioned, this phenomenon is described by Elman [2][3].

After level 1 has completed its training, the test program turns off learning in level 1 and enables learning in level 2. In contrast to level 1, which learns to recognize sequences of characters as words, level 2 learns to recognize sequences of words as phrases. During the training phase for level 2, the test program sends an input stream of events to level 1, and level 1, in turn, generates the input stream of events to level 2. The input stream generated by the test program consists of the characters in phrases that are made up of the words that have already been learned by level 1.

In the following test program, the programmer has somewhat arbitrarily partitioned the words in the Gettysburg Address into a collection of phrases. The program selects phrases at random from this collection and appends the characters in these phrases to the input stream to level 1:

```c
char *test2[] =
{
    "fourscoreandsevenyearsago",
    "ourfathersbroughtforthonthiscontinent",
    "anewnation",
    "conceivedinliberty",
    "anddedicatedtotheproposition",
    "thatallmen",
    "arecreatedequal",
    "nowweareengagedinagreatcivilwar",
    "testingwhetherthatnation",
    "oranynationsoconceivedandsodedicated",
    "canlongendure",
    "wearemetonagreatbattlefieldofthatwar",
    "wehavecometodedicateaportionofthatfield",
```

```
        "asafinalrestingplace",
        "forthosewhoheregavetheirlives",
        "thatthatnationmightlive",
        "itisaltogetherfitting",
        "andproper",
        "thatweshoulddothis",
        "butinalargersense",
        "wecannotdedicate",
        "wecannotconsecrate",
        "wecannothallowthisground",
        "thebravemen",
        "livinganddead",
        "whostruggledhere",
        "haveconsecratedit",
        "faraboveourpoorpowertoaddordetract",
        "theworldwilllittlenote",
        "norlongremember",

        ...

        "shallnotperishfromtheearth",
};
const int NUM_TEST_PHRASES = sizeof(test1)/sizeof(test1[0]);
const int NUM_LEARNING_PHRASES = 150000;

void main()
{
    for (int ix = 0; ix < NUM_LEARNING_PHRASES; ++ix)
    {
        int jx = rand() % NUM_TEST_PHRASES;
        printf("%s", test2[jx]);
    }
}
```

After level 2 is trained, it can do the following:

- Identify the boundaries between successive phrases in the input stream.
- Make local predictions about what the next word in the input stream will be.

The test program again sends the following stream of characters to level 1:

```
fourscoreandsevenyearsagoourfathersbroughtforthonthiscontinentanewnationconce
ivedinlibertyanddedicatedtothepropositionthatallmenarecreatedequal...
```

Now that both levels have finished their training, they can jointly convert this input stream into an output stream in which level 2 marks the boundaries between phrases in this stream by inserting newline characters into the stream, and level 1 marks the word boundaries by inserting spaces, as follows:

```
four score and seven years ago
our fathers brought forth on this continent
a new nation
conceived in liberty
and dedicated to the proposition
that all men
are created equal
now we are engaged in a great civil war
```

```
testing whether that nation
or any nation so conceived and so dedicated
can long endure
we are met on a great battlefield of that war
we have come to dedicate a portion of that field
as a final resting place
for those who here gave their lives
that that nation might live
it is altogether fitting
and proper
that we should do this...
```

By incorporating feedback from level 2, level 1 does a much better job of predicting the next character to arrive in its input stream. As before, the test program compares the actual next character with the prediction from level 1 and highlights each *incorrectly* predicted character by converting the character from lower case to upper case:

```
FoUrscoreandsevenyearsagoOUrfathersbroughtforthonthiscontinentANEwNationCOnCe
ivedinlibertyANddedicatedtothepropositionThAtAllmenAReCreatedequalNOWweareeng
agedinagreatcivilwarTEstingwhetherthatnationORanynationsoconceivedandsodedica
tedCanLongendureWeAreMetonagreatbattlefieldofthatwarWeHavecometodedicateaport
ionofthatfieldASafinalrestingplaceForThosewhoheregavetheirlivesThAtThAtnation
mightliveItisaltogetherfittingANdProperThAtweshoulddothisBUtiNalargersenseWec
annotdedicateWecannotConsecrateWecannotHallowthisgroundTheBravemenLivingandde
adWHoStruggledhereHAveConsecrateditFArAboveourpoorpowertoaddordetractTheWorld
willlittlenoteNORlongrememberWHAtwesayhereBUtitcanneverforgetwhattheydidhereI
tisForusTheLivingRathertobededicatedhereTOtheUnfinishedworkWHIchtheyWhofought
hereHAvethusfarsonoblyadvancedItisRatherforusTOBeherededicatedTOtheGreattaskr
emainingbeforeusThAtFRomthesehonoreddeadWeTakeincreaseddevotionTOthAtcauseFor
WhichtheygaveTheLAstfullmeasureofdevotionThAtweHerehighlyresolveThAtThesedead
SHallnothavediedinvain...
```

Most of the prediction errors now occur in the first few characters at the start of a new phrase. With help from level 2, level 1 now does a good job of predicting the remaining characters in each phrase.

## Discussion

This paper has presented a high-level model of the interaction between two adjacent levels in a cortical hierarchy. This model is a simplified version of the memory-prediction framework proposed by Hawkins and Blakeslee [4], and implements only the features required to model the role of cortical feedback in making predictions about future events.

Each level in the hierarchy tries to predict the next event in its input stream before the event occurs. The success of such predictions is greatly improved if solid predictions are available from the higher levels in the hierarchy.

No attempt is made at any level in the hierarchy to predict more than one event into the future. However, the predictions made at higher levels encompass events that are increasingly distant in time compared with the predictions made at lower levels. For example, if level 1—at the bottom of the hierarchy—tries to predict the next *character* in its input stream, then level 2—the next higher level—tries to predict the next *word*, level 3 tries to predict the next *phrase*, and so on.

# Appendix: Program Listing

The C++ source code listing in this appendix is an implementation of the high-level neocortical model described in this paper, plus a test program to exercise the model. The results described in this paper are based on this code. The test results are printed as text to the standard output device. To run the program, copy all of the following source code into a file that has a .cpp file name extension. Then compile the program and run it.

```cpp
//
// Use the event window segmentation (EWS) algorithm
// to experiment with feedback in a high-level model
// of a neocortical hierarchy that has two levels.
//

#pragma warning(disable:4996)
#define SHOW_DIAGNOSTICS  0

#include <stdlib.h>
#include <string.h>
#include <memory.h>
#include <stdio.h>
#include <assert.h>
#include <math.h>

#define min(x,y)   ((x)<(y)?(x):(y))
#define max(x,y)   ((x)>(y)?(x):(y))
#define ARRAY_LENGTH(x)  (sizeof(x)/sizeof(x[0]))

const int MAX_ALPHABET_SYMBOLS = 256;
const int MAX_WINDOW_WIDTH = 32;
const float THRESHOLD_BIAS = 4.567;
const float HIGHLY_LIKELY = 0.99;
const float HIGHLY_UNLIKELY = (1.0 - HIGHLY_LIKELY);
const int END_OF_STREAM = 0xff;

typedef unsigned char UINT8;
typedef unsigned short UINT16;
typedef unsigned long UINT32;

// A sequence stored in sequence memory
typedef struct _SEQUENCE
{
    struct _SEQUENCE *_link1;
    struct _SEQUENCE *_link2;
    struct _SEQUENCE *_prevSeq;
    struct
    {
        UINT8 _event;
        UINT8 _length;
        UINT16 _allocNum;
    } _info;
    UINT32 _createCount;
    UINT32 _inCount;
    UINT32 _outCount;
    UINT32 _succCount;
    float _accumScores;
    UINT8 _sequenceName;
    UINT8 _precursorCount;
} SEQUENCE;

// A cell in a column in an event window
typedef struct _CELL
{
    SEQUENCE *_seq;
    float _score;
} CELL;

// A column in an event window
typedef struct _COLUMN
{
    struct _COLUMN *_prevCol;
    struct _COLUMN *_nextCol;
    UINT8 _event;
    float _bestScore;
    int _bestLength;
    int _numCells;
    CELL _cell[MAX_WINDOW_WIDTH];
} COLUMN;

//
// Test data and tuned parameters
//
#if 0

const int NUM_LEARNING_WORDS = 500;
const float THRESHOLD_PROB = 0.4 / 44.0;
```

```c
char *_test1[] =     // word = sequence of letters
{
    "the", "quick", "brown", "fox", "jumped",
    "over", "the", "lazy", "sleeping", "dog"
};

char *_test2[] =     // phrase = sequence of words
{
    "thequickbrownfoxjumped",
    "overthelazysleepingdog"
};

#elif 1

const int NUM_LEARNING_WORDS = 150000;
const float THRESHOLD_PROB = 0.76 / 1149.0;

char *_test1[] =     // word = sequence of letters
{
    "four", "score", "and", "seven", "years", "ago", "our", "fathers",
    "brought", "forth", "on", "this", "continent", "a", "new",
    "nation", "conceived", "in", "liberty", "and", "dedicated", "to",
    "the", "proposition", "that", "all", "men", "are", "created",
    "equal", "now", "we", "are", "engaged", "in", "a", "great",
    "civil", "war", "testing", "whether", "that", "nation", "or",
    "any", "nation", "so", "conceived", "and", "so", "dedicated",
    "can", "long", "endure", "we", "are", "met", "on", "a", "great",
    "battlefield", "of", "that", "war", "we", "have", "come", "to",
    "dedicate", "a", "portion", "of", "that", "field", "as", "a",
    "final", "resting", "place", "for", "those", "who", "here",
    "gave", "their", "lives", "that", "that", "nation", "might",
    "live", "it", "is", "altogether", "fitting", "and", "proper",
    "that", "we", "should", "do", "this", "but", "in", "a", "larger",
    "sense", "we", "can", "not", "dedicate", "we", "can", "not",
    "consecrate", "we", "can", "not", "hallow", "this", "ground",
    "the", "brave", "men", "living", "and", "dead", "who",
    "struggled", "here", "have", "consecrated", "it", "far", "above",
    "our", "poor", "power", "to", "add", "or", "detract", "the",
    "world", "will", "little", "note", "nor", "long", "remember",
    "what", "we", "say", "here", "but", "it", "can", "never",
    "forget", "what", "they", "did", "here", "it", "is", "for", "us",
    "the", "living", "rather", "to", "be", "dedicated", "here", "to",
    "the", "unfinished", "work", "which", "they", "who", "fought",
    "here", "have", "thus", "far", "so", "nobly", "advanced", "it",
    "is", "rather", "for", "us", "to", "be", "here", "dedicated",
    "to", "the", "great", "task", "remaining", "before", "us", "that",
    "from", "these", "honored", "dead", "we", "take", "increased",
    "devotion", "to", "that", "cause", "for", "which", "they", "gave",
    "the", "last", "full", "measure", "of", "devotion", "that", "we",
    "here", "highly", "resolve", "that", "these", "dead", "shall",
    "not", "have", "died", "in", "vain", "that", "this", "nation",
    "under", "god", "shall", "have", "a", "new", "birth", "of",
    "freedom", "and", "that", "government", "of", "the", "people",
    "by", "the", "people", "for", "the", "people", "shall", "not",
    "perish", "from", "the", "earth"
};

char *_test2[] =     // phrase = sequence of words
{
    "fourscoreandsevenyearsago",
    "ourfathersbroughtforthonthiscontinent",
    "anewnation",
    "conceivedinliberty",
    "anddedicatedtotheproposition",
    "thatallmen",
    "arecreatedequal",
    "nowweareengagedinagreatcivilwar",
    "testingwhetherthatnation",
    "oranynationsoconceivedandsodedicated",
    "canlongendure",
    "wearemetonagreatbattlefieldofthatwar",
    "wehavecometodedicateaportionofthatfield",
    "asafinalrestingplace",
    "forthosewhoheregavetheirlives",
    "thatthatnationmightlive",
    "itisaltogetherfitting",
    "andproper",
    "thatweshoulddothis",
    "butinalargersense",
    "wecannotdedicate",
    "wecannotconsecrate",
    "wecannothallowthisground",
    "thebravemen",
    "livinganddead",
    "whostruggledhere",
    "haveconsecratedit",
    "faraboveourpoorpowertoaddordetract",
    "theworldwilllittlenote",
    "norlongremember",
    "whatwesayhere",
    "butitcanneverforgetwhattheydidhere",
    "itisforus",
    "theliving",
    "rathertobededicatedhere",
    "totheunfinishedwork",
```

```cpp
        "whichtheywhofoughthere",
        "havethusfarsonoblyadvanced",
        "itisratherforus",
        "tobeherededicated",
        "tothegreattaskremainingbeforeus",
        "thatfromthesehonoreddead",
        "wetakeincreaseddevotion",
        "tothatcause",
        "forwhichtheygave",
        "thelastfullmeasureofdevotion",
        "thatwewherehighlyresolve",
        "thatthesedead",
        "shallnothavediedinvain",
        "thatthisnation",
        "undergod",
        "shallhaveanewbirthoffreedom",
        "andthatgovernmentofthepeople",
        "bythepeople",
        "forthepeople",
        "shallnotperishfromtheearth",
};

#endif

//
// A bitmask to represent a set of up to 255 items. If bit N is one, item N
// is a member of the set. Note that bit 0 is reserved and is always zero.
//
class Mask255
{
    // Return bit number of rightmost one in val.
    int _rmo(UINT32 val)
    {
        int n, m = 31;

        assert(val);
        for (n = 16; n; n >>= 1)
            if (val << n)
            {
                val <<= n;
                m -= n;
            }

        return m;
    }
public:
    UINT32 _mask32[8];

    Mask255()
    {
        ClearMask();
    }

    void ClearMask()
    {
        memset(_mask32, 0, sizeof(_mask32));
    }

    void AndMask(Mask255 *maskIn)
    {
        for (int ix = 0; ix < ARRAY_LENGTH(_mask32); ++ix)
            _mask32[ix] &= maskIn->_mask32[ix];
    }

    void OrMask(Mask255 *maskIn)
    {
        for (int ix = 0; ix < ARRAY_LENGTH(_mask32); ++ix)
            _mask32[ix] |= maskIn->_mask32[ix];
    }

    void SetBit(int index)
    {
        assert(0 < index && index < 8 * sizeof(_mask32));
        _mask32[index >> 5] |= 1 << (index & 31);
    }

    void ClearBit(int index)
    {
        assert(0 < index && index < 8 * sizeof(_mask32));
        _mask32[index >> 5] &= ~(1 << (index & 31));
    }

    int GetBit(int index)
    {
        assert(0 < index && index < 8 * sizeof(_mask32));
        int val = _mask32[index >> 5] >> (index & 31);
        return (val & 1);
    }

    // Given number of previous bit, return number of next bit in mask.
    int GetNext(int prev)
    {
        int ixHigh = prev >> 5;
        int ixLow = prev & 31;
```

```cpp
            assert(0 <= prev && prev < 8 * sizeof(_mask32));
            if (ixLow != 31)
            {
                int mask = (~0) << (ixLow + 1);
                int val = _mask32[ixHigh] & mask;

                if (val)
                    return (32 * ixHigh + _rmo(val));
            }
            for (++ixHigh; ixHigh < ARRAY_LENGTH(_mask32); ++ixHigh)
                if (_mask32[ixHigh])
                    return (32 * ixHigh + _rmo(_mask32[ixHigh]));

            // There is no next bit set in the 255-bit mask.
            return 0;
        }

        // Return true if mask is all zeros. Otherwise, return false.
        bool IsEmpty()
        {
            for (int ix = 0; ix < ARRAY_LENGTH(_mask32); ++ix)
                if (_mask32[ix])
                    return false;

            return true;
        }
};

//
// A virtual class that represents a region in a hierarchy of regions
//
class Region
{
public:
    virtual void ProcessEvents(UINT8 *evt, int hardCount) = 0;
    virtual UINT8 RetractLastEvent() = 0;
    virtual void ConnectToRegionAbove(Region *region) = 0;
    virtual void WriteString(UINT8 *string) = 0;
    virtual void ApplyFeedbackFilter(Mask255 *mask) = 0;
    virtual void SetFeedback(UINT8 name, float score) = 0;
};

//
// A class that represents the top region in a hierarchy of regions.
// This region is minimally functional and mostly just logs results.
//
class TopRegion : public Region
{
    Region *_regionBelow;

public:
    TopRegion() : _regionBelow(NULL) { }

    ~TopRegion() { }

    void ProcessEvents(UINT8 *evt, int hardCount)
    {
        UINT8 str[2];

        str[1] = 0;
        for (int ix = 0; ix < hardCount; ++ix)
        {
            str[0] = evt[ix];
            WriteString(str);
        }
    }

    UINT8 RetractLastEvent()
    {
        return 0;
    }

    void ConnectToRegionAbove(Region *above)
    {
        assert(!above);
    }

    void ConnectToRegionBelow(Region *below)
    {
        _regionBelow = below;
        if (_regionBelow)
            _regionBelow->ConnectToRegionAbove(this);
    }

    void WriteString(UINT8 *string)
    {
#if SHOW_DIAGNOSTICS != 0
        printf("\n\n  .   .   .   ");
        assert(_regionBelow);
        _regionBelow->WriteString(string);
#endif
    }

    void ApplyFeedbackFilter(Mask255 *mask)
```

```cpp
        {
        }

        void SetFeedback(UINT8 name, float score)
        {
        }
};

//
// A class that represents the bottom region in a hierarchy of regions.
// This region is minimally functional and mostly just logs results.
//
class BottomRegion : public Region
{
    UINT8 _predictedEvent;
    float _predictionScore;
    Region *_regionAbove;

public:
    BottomRegion() :
        _predictedEvent(0),
        _predictionScore(0.0),
        _regionAbove(NULL)
    { }

    ~BottomRegion() { }

    void ProcessEvents(UINT8 *evt, int hardCount)
    {
        assert(0);
    }

    UINT8 RetractLastEvent()
    {
        assert(0);
        return 0;
    }

    void ConnectToRegionAbove(Region *above)
    {
        assert(above);
        _regionAbove = above;
    }

    void WriteString(UINT8 *string)
    {
        if (string[0] == END_OF_STREAM)
        {
            assert(string[1] == '\0');
            printf("<end of stream>");
        }
        else
            printf("%s", string);
    }

    void ApplyFeedbackFilter(Mask255 *mask)
    {
    }

    void SetFeedback(UINT8 name, float score)
    {
        _predictedEvent = name;
        _predictionScore = score;
    }

    // Query region above for its most recent prediction.
    UINT8 GetFeedback()
    {
        _regionAbove->ApplyFeedbackFilter(NULL);
        return _predictedEvent;
    }
};

//
// A Trainer speeds up training so that experiments run more quickly.
// A trainer is sandwiched between the region below, which has completed
// its training, and the region above, which needs to be trained. The
// trainer captures a full set of sequences from the region below, and
// then plays back these sequences many times and in random order to the
// region above. Using a Trainer is much faster than propagating events
// from the bottom of the hierarchy, but produces very similar results.
// After the region above finishes its training, the trainer is removed.
//
class Trainer : public Region
{
    UINT8 _data[1024], *_word[256];  // BUGBUG - Static allocation
    Region *_regionAbove, *_regionBelow;
    int _numData, _numWords;

public:
    Trainer() :
        _regionAbove(NULL),
        _regionBelow(NULL),
        _numData(0),
        _numWords(0)
```

```cpp
    {
        memset(&_data[0], 0, sizeof(_data));
        memset(&_word[0], 0, sizeof(_word));
    }

    ~Trainer() { }

    // Capture all hard events from region below in a data buffer.
    void ProcessEvents(UINT8 *evt, int hardCount)
    {
        assert(_numData + hardCount < ARRAY_LENGTH(_data));
        memcpy(&_data[_numData], &evt[0], hardCount);
        _numData += hardCount;
    }

    UINT8 RetractLastEvent()
    {
        return 0;
    }

    // Process stored event stream to find end-of-sequence markers.
    // Then play back sequences many times and in random order.
    void ConnectToRegionAbove(Region *above)
    {
        int ix;
        UINT8 buf[2];

        assert(above);
        _regionAbove = above;
        buf[1] = '\0';
        _word[0] = &_data[0];
        _numWords = 0;
        for (ix = 0; ix < _numData; ++ix)
        {
            if (_data[ix] == END_OF_STREAM)  // Break marker
            {
                _data[ix] = '\0';   // Null-terminated string
                _word[++_numWords] = &_data[ix+1];
            }
        }

        // For easier debugging, make sequence names predictable.
        for (ix = 0; ix < _numWords; ++ix)
        {
            for (UINT8 *s = _word[ix]; *s; ++s)
            {
                buf[0] = *s;
                _regionAbove->ProcessEvents((UINT8*)buf, 1);
            }
        }

        // BUGBUG - Same NUM_LEARNING_WORDS for all regions?
        for (ix = 0; ix < NUM_LEARNING_WORDS; ++ix)
        {
            int index = rand() % _numWords;

            for (UINT8 *s = _word[index]; *s; ++s)
            {
                buf[0] = *s;
                _regionAbove->ProcessEvents((UINT8*)buf, 1);
            }
        }
    }

    void ConnectToRegionBelow(Region *below)
    {
        _regionBelow = below;
        if (_regionBelow)
            _regionBelow->ConnectToRegionAbove(this);
    }

    void WriteString(UINT8 *string)
    {
        printf("\n .  .  .  ");
        assert(_regionBelow);
        _regionBelow->WriteString(string);
    }

    void ApplyFeedbackFilter(Mask255 *mask)
    {
    }

    void SetFeedback(UINT8 name, float score)
    {
    }
};

//
// A class that represents a middle region in a hierarchy of regions.
// This region learns recurring sequences of events in input stream.
//
class MidRegion : public Region
{
public:
    int _instance;
```

```cpp
    Region *_regionBelow;
    Region *_regionAbove;
    char _sequenceDelimiter[8];
    SEQUENCE _rootSequence[MAX_ALPHABET_SYMBOLS];
    SEQUENCE *_namedSequence[MAX_ALPHABET_SYMBOLS];
    SEQUENCE *_hashTable1[12577];
    SEQUENCE *_hashTable2[1259];
    COLUMN _localPredictions;
    COLUMN _column[MAX_WINDOW_WIDTH];
    COLUMN *_headColumn;
    int _numColumns;
    int _numSoftColumns;
    int _minColumns;
    int _allocCount;
    int _eventCount;
    int _fireCount;
    int _nameIndex;
    bool _enablePredictions;
    int _outputOffset;
    CELL _predictionFromAbove;
    bool _feedbackSourceIsLocal;
    Mask255 *_eventSet[4][MAX_ALPHABET_SYMBOLS];
    Mask255 _feedbackFilter;
    int _retractedName;

    MidRegion(char *delim);
    ~MidRegion();
    void ProcessEvents(UINT8 *evt, int hardCount);
    void ConnectToRegionAbove(Region *above)
    {
        _regionAbove = above;
    }
    void ConnectToRegionBelow(Region *below)
    {
        _regionBelow = below;
        if (_regionBelow)
            _regionBelow->ConnectToRegionAbove(this);
    }
    void WriteString(UINT8 *string);
    void PruneSequenceMemory();
    void BuildSuccessorLists();
    void InitLocalPredictions();
    void InitFeedbackFiltering();
    bool BuildFeedbackFilter(Mask255 *mask);
    void ApplyFeedbackFilter(Mask255 *mask);
    void SetFeedback(UINT8 name, float prob);
    void DumpSequenceMemory();
    void DumpNamedSequences();

private:
    static int _nextInstance;
    UINT32 GetHashIndex(SEQUENCE *seq, UINT8 event);
    SEQUENCE* NextSequence(SEQUENCE *seq, UINT8 event);
    void PopulateNewColumn(UINT8 event, COLUMN *inCol, COLUMN *outCol);
    void SetFirstLevelScores();
    void SetSecondLevelScores();
    bool EmitWords();
    bool IsSegmentable(int offset, COLUMN *col);
    UINT8 RetractLastEvent();
};

int MidRegion::_nextInstance = 1;

//
// Initialize data structures used by the word-segmentation algorithm.
//
MidRegion::MidRegion(char *delim) :
    _regionBelow(NULL),
    _regionAbove(NULL),
    _headColumn(NULL),
    _numColumns(0),
    _numSoftColumns(0),
    _minColumns(MAX_WINDOW_WIDTH/2),
    _allocCount(0),
    _eventCount(0),
    _fireCount(0),
    _nameIndex(0),
    _enablePredictions(false),
    _outputOffset(0),
    _feedbackSourceIsLocal(false),
    _retractedName(0)
{
    int ix;

    _instance = MidRegion::_nextInstance++;
    memset(_hashTable1, 0, sizeof(_hashTable1));
    memset(_hashTable1, 0, sizeof(_hashTable2));
    memset(&_localPredictions, 0, sizeof(_localPredictions));
    memset(&_predictionFromAbove, 0, sizeof(_predictionFromAbove));
    memset(_eventSet, 0, sizeof(_eventSet));
    memset(_rootSequence, 0, sizeof(_rootSequence));
    for (ix = 1; ix < ARRAY_LENGTH(_rootSequence); ++ix)
    {
        _rootSequence[ix]._info._event = ix;
        _rootSequence[ix]._info._length = 1;
```

```cpp
            _rootSequence[ix]._info._allocNum = ++_allocCount;
            assert(_rootSequence[ix]._accumScores == 0.0);
        }
        _rootSequence[END_OF_STREAM]._info._event = END_OF_STREAM;
        _rootSequence[END_OF_STREAM]._sequenceName = END_OF_STREAM;
        memset(_namedSequence, 0, sizeof(_namedSequence));
        _namedSequence[END_OF_STREAM] = &_rootSequence[END_OF_STREAM];
        memset(_column, 0, sizeof(_column));
        for (ix = 1; ix < MAX_WINDOW_WIDTH; ++ix)
        {
            _column[ix-1]._nextCol = &_column[ix];
            _column[ix]._prevCol = &_column[ix-1];
        }
        _column[MAX_WINDOW_WIDTH - 1]._nextCol = &_column[0];
        _column[0]._prevCol = &_column[MAX_WINDOW_WIDTH - 1];
        memset(_sequenceDelimiter, 0, sizeof(_sequenceDelimiter));
        assert(strlen(delim) < ARRAY_LENGTH(_sequenceDelimiter));
        strcpy(_sequenceDelimiter, delim);
}

//
// Free data structures allocated by the event window segmentation (EWS) algorithm.
//
MidRegion::~MidRegion()
{
    for (int ix = 0; ix < ARRAY_LENGTH(_hashTable1); ++ix)
    {
        SEQUENCE *seq = _hashTable1[ix];

        while (seq)
        {
            SEQUENCE *temp = seq;

            seq = seq->_link1;
            delete temp;
        }
    }
}

//
// Use the ELF hash algorithm to generate a hash table index.
//
UINT32 MidRegion::GetHashIndex(SEQUENCE *seq, UINT8 event)
{
    const UINT32 count = sizeof(event) + sizeof(seq->_info);
    UINT8 *p = (UINT8*)(&seq->_info);
    UINT32 h = 0, g;

    for (int ix = 0; ix < count; ++ix)
    {
        h = (h << 4) + event;
        if (g = h & 0xF0000000)
            h ^= g >> 24;

        h &= ~g;
        event = *p++;
    }
    return h;
}

//
// Look up the sequence that is formed by appending a character
// to the specified input sequence.
//
SEQUENCE* MidRegion::NextSequence(SEQUENCE *seq, UINT8 event)
{
    SEQUENCE *p, *q;
    int index;

    if (!seq)
        return &_rootSequence[event];

    index = GetHashIndex(seq, event) % ARRAY_LENGTH(_hashTable1);
    for (p = _hashTable1[index], q = NULL; p; q = p, p = p->_link1)
        if (p->_prevSeq == seq && p->_info._event == event)
        {
            if (!q)
                return p;

            q->_link1 = p->_link1;
            break;
        }

    if (!p)
    {
        p = new SEQUENCE;
        assert(p);
        memset(p, 0, sizeof(SEQUENCE));
        p->_prevSeq = seq;
        p->_info._event = event;
        p->_info._length = 1 + seq->_info._length;
        p->_info._allocNum = ++_allocCount;
        p->_createCount = _eventCount;
        assert(p->_accumScores == 0.0);
    }
```

```cpp
        p->_link1 = _hashTable1[index];
        _hashTable1[index] = p;
        return p;
}

//
// Populate a new column in the event window with pointers to valid
// sequences that match the latest characters to arrive in the input
// stream. Additionally, update the statistics in the sequence memory.
//
void MidRegion::PopulateNewColumn(UINT8 event, COLUMN *inCol, COLUMN *outCol)
{
    int ix, incr = (!_regionAbove) ? 1 : 0;  // In learning mode?
    int numInCells = (!inCol) ? 0 : inCol->_numCells;
    CELL *inCell = (!inCol) ? NULL : &inCol->_cell[0];
    CELL *outCell = &outCol->_cell[0];
    SEQUENCE *inSeq, *outSeq;

    assert(event);
    assert(outCol);
    _eventCount += incr;

    memset(outCol->_cell, 0, sizeof(outCol->_cell));
    outCol->_event = event;
    outCol->_bestLength = -1;
    outCol->_bestScore = 0.0;

    outSeq = NextSequence(NULL, event);
    outSeq->_inCount += incr;
    outSeq->_outCount += incr;
    outCell[0]._seq = outSeq;
    assert(outCell[0]._score == 0.0);
    outCol->_numCells = 1;
    for (ix = 0; ix < numInCells; ++ix)
        {
        float count, prob;

        inSeq = inCell[ix]._seq;
        outSeq = NextSequence(inSeq, event);
        outSeq->_inCount += incr;
        count = _eventCount - outSeq->_createCount;
        prob = (outSeq->_inCount - THRESHOLD_BIAS) / count;
        if (prob < THRESHOLD_PROB)
            break;

        if (!outSeq->_outCount)
            ++_fireCount;

        outSeq->_outCount += incr;
        inSeq->_succCount += incr;
        outCell[outCol->_numCells]._seq = outSeq;
        outCol->_numCells += 1;
        assert(outCol->_numCells < MAX_WINDOW_WIDTH);
        assert(outSeq->_info._length == outCol->_numCells);
        }

    // Update the statistics on these mostly junk sequences
    // because a few of them will turn out to be valid sequences.
    for (++ix; ix < numInCells; ++ix)
        {
        inSeq = inCell[ix]._seq;
        outSeq = NextSequence(inSeq, event);
        outSeq->_inCount += incr;
        }
}

//
// Calculate first-level scores for the valid sequences in the new column.
//
void MidRegion::SetFirstLevelScores()
{
    COLUMN *col = _headColumn->_prevCol;
    CELL *inCell = col->_cell;
    CELL *outCell = _headColumn->_cell;
    float Psow = (float)outCell[0]._seq->_outCount / _eventCount;
    float sum = 0.0;

    assert(_headColumn->_numCells > 0);
    for (int ix = 0; ix < col->_numCells; ++ix)
        {
        SEQUENCE *inSeq = inCell[ix]._seq;
        float score = inCell[ix]._score;
        float eowCount = inSeq->_outCount - inSeq->_succCount;
        float Peow = eowCount / inSeq->_outCount;
        float Pnew = Peow * Psow;

        if (ix + 1 < _headColumn->_numCells)
            {
            SEQUENCE *outSeq = outCell[ix+1]._seq;
            float Psame = (float)outSeq->_outCount / inSeq->_outCount - Pnew;

            outCell[ix+1]._score = score * Psame;
            sum += outCell[ix+1]._score;
            }
        score *= Pnew;
```

```cpp
            if (!ix || outCell[0]._score < score)
            {
                outCell[0]._score = score;
                col->_bestLength = ix + 1;
            }
        }

        // Normalize first-level scores in new column.
        if (sum != 0.0)
        {
            sum += outCell[0]._score;
            for (int jx = 0; jx < _headColumn->_numCells; ++jx)
                outCell[jx]._score /= sum;
        }
        else
            outCell[0]._score = 1.0;

        // Second-level score will use average first-level score from
        // survivor path to start of hypothetical new word.
        if (!_regionAbove)
            inCell[col->_bestLength-1]._seq->_accumScores += outCell[0]._score;
}

//
// Calculate second-level scores for the paths in the event window.
//
void MidRegion::SetSecondLevelScores()
{
    COLUMN *col, *prevCol;

    assert(_numColumns);
    prevCol = _headColumn - _numColumns;
    if (prevCol < _column)
        prevCol += ARRAY_LENGTH(_column);

    prevCol->_bestScore = 1.0;
    col = prevCol->_nextCol;
    assert(col->_numCells == 1);
    for ( ; col != _headColumn; prevCol = col, col = col->_nextCol)
    {
        for (int ix = 0; ix < col->_numCells; ++ix)
        {
            SEQUENCE *seq = col->_cell[ix]._seq;
            float Pword = seq->_accumScores / seq->_inCount;
            float boost = 1.0 + 2.0/(ix + 1);
            float score = prevCol->_bestScore;

            score *= pow(Pword, ix + 1) * boost;
            if (!ix || col->_bestScore < score)
            {
                col->_bestScore = score;
                col->_bestLength = ix + 1;
            }
            prevCol = prevCol->_prevCol;
        }
    }
}

//
// Emit zero, one, or more words to the region above. Some number
// of words emitted from the left side of the event window might
// be permanently detached from the window. Emitted words that
// are not permanently detached might change in future emissions.
// Return 1 if any words were emitted; otherwise, return 0.
//
bool MidRegion::EmitWords()
{
    COLUMN *col;
    UINT8 buffer[MAX_WINDOW_WIDTH];
    UINT8 *str = &buffer[ARRAY_LENGTH(buffer)];
    int len, hardCount, hardLength, minOffset = max(_minColumns/2, _numSoftColumns);

    // Identify event window path that contains words to emit.
    assert(_headColumn);
    assert(_numColumns <= MAX_WINDOW_WIDTH);
    if (_headColumn->_event == END_OF_STREAM)
    {
        // Flush event window contents.
        minOffset = -1;
        _headColumn->_bestLength = 1;
        _outputOffset = 0;
    }
    else
    {
        _headColumn->_bestScore = 0.0;
        if (_predictionFromAbove._score < 0.5)
        {
            CELL *cell = &_headColumn->_cell[0];

            // Find event window path with highest first-level score.
            for (int ix = 0; ix < _headColumn->_numCells; ++ix)
            {
                if (!ix || _headColumn->_bestScore < cell[ix]._score)
                {
                    _headColumn->_bestScore = cell[ix]._score;
```

```
                    _headColumn->_bestLength = ix + 1;
                }
            }
        }
        if (_predictionFromAbove._score > _headColumn->_bestScore)
        {
            // Use prediction from above to identify event window path.
            assert(_predictionFromAbove._seq);
            assert(_outputOffset);
            len = _predictionFromAbove._seq->_info._length;
            if (len <= _outputOffset)
            {
                _outputOffset -= len;
                col = _headColumn - _outputOffset;
                if (col < _column)
                    col += ARRAY_LENGTH(_column);

                assert(len <= col->_numCells);
                col->_bestLength = len;
            }
            else
                _headColumn->_bestLength = _outputOffset;

            if (_predictionFromAbove._score > HIGHLY_LIKELY)
                minOffset = -1;  // Detach _all_ words in event window.
        }
        else
            _outputOffset = _headColumn->_bestLength;
    }
    assert(0 <= _outputOffset);
    assert(_outputOffset <= ARRAY_LENGTH(_column));

    // Gather list of all words in highest-scoring path.
    hardCount = 0;
    hardLength = 0;
    *--str = '\0';
    len = 0;
    for (int offset = _outputOffset; offset < _numColumns; offset += len)
    {
        SEQUENCE *seq;

        col = _headColumn - offset;
        if (col < _column)
            col += ARRAY_LENGTH(_column);

        assert(col >= &_column[0]);
        len = col->_bestLength;
        seq = col->_cell[len-1]._seq;
        if (!seq->_sequenceName)
        {
            // This sequence has never been used as an output word
            // before now, so it needs to be assigned a name.
            seq->_sequenceName = ++_nameIndex;
            assert(_nameIndex < ARRAY_LENGTH(_namedSequence));
            _namedSequence[_nameIndex] = seq;
            for (SEQUENCE *seq2 = seq; seq2; seq2 = seq2->_prevSeq)
                seq2->_precursorCount += 1;
        }
        *--str = seq->_sequenceName;
        if (offset > minOffset)
        {
            // Count detached word and accumulate word length.
            ++hardCount;
            hardLength += seq->_info._length;
        }
    }
    if (!str[0])
    {
        assert(_numColumns < MAX_WINDOW_WIDTH);  // Buffer overflow?
        return false;
    }
    if (_regionAbove)
        _regionAbove->ProcessEvents(str, hardCount);

    _numColumns -= hardLength;
    assert(_numColumns >= 0);
    if (_numColumns)
    {
        // Align left edge of event window to new word boundary.
        col = _headColumn - _numColumns + 1;
        if (col < _column)
            col += ARRAY_LENGTH(_column);

        for (len = 1;
             len <= _numColumns && col->_numCells > len;
             ++len, col = col->_nextCol)
        {
            col->_numCells = len;
            if (col->_bestLength > len)
                col->_bestLength = len;
        }
    }
    else
        _headColumn = NULL;  // Event window is empty.
```

```cpp
        return true;
}

//
// Process the latest events to arrive in the input stream.
//
void MidRegion::ProcessEvents(UINT8 *evt, int hardCount)
{
    COLUMN *col;
    SEQUENCE *seq;
    int ix, len;
    UINT8 event;

    assert(evt && evt[0]);
    assert(hardCount <= strlen((char*)evt));
    _localPredictions._bestLength = 0;

    // Skip past events that match what's already in the event window.
    if (_numSoftColumns)
    {
        int numHardColumns = _numColumns - _numSoftColumns;

        assert(_regionAbove);
        assert(_headColumn);
        col = _headColumn - _numSoftColumns + 1;
        if (col < _column)
            col += ARRAY_LENGTH(_column);

        for (ix = 0; ix < _numSoftColumns; ++ix, col = col->_nextCol)
            if (col->_event != evt[ix])
                break;

        if (ix < _numSoftColumns)
        {
            _numColumns = ix + numHardColumns;
            _headColumn = (_numColumns) ? col->_prevCol : NULL;
        }
        if (hardCount > ix)
        {
            assert(evt[ix]);
            numHardColumns += ix;
            hardCount -= ix;
        }
        else
        {
            numHardColumns += hardCount;
            hardCount = 0;
        }
        _numSoftColumns = _numColumns - numHardColumns;
        if (!evt[ix])
            return;  // No new events to process.

        evt = &evt[ix];
    }

    // Add initial event to empty, uninitialized event window.
    if (!_headColumn)
    {
        if (!_regionAbove && !hardCount)
            return;

        event = *evt++;
        assert(event);
        _headColumn = &_column[0];
        PopulateNewColumn(event, NULL, _headColumn);
        _outputOffset = 1;
        _headColumn->_cell[0]._score = 1.0;
        if (hardCount)
        {
            --hardCount;
            assert(!_numSoftColumns);
        }
        else
            _numSoftColumns = 1;

        if (_regionAbove)
            _numColumns = 1;
    }

    // Include soft events only if there's a region above.
    len = (_regionAbove) ? strlen((char*)evt) : hardCount;

    // Add new events to nonempty, initialized event window.
    for (ix = 0; ix < len; ++ix)
    {
        event = *evt++;
        assert(event);
        col = _headColumn;
        _headColumn = _headColumn->_nextCol;
        PopulateNewColumn(event, col, _headColumn);
        assert(_headColumn->_numCells < MAX_WINDOW_WIDTH);
        ++_outputOffset;
        SetFirstLevelScores();
        if (!hardCount)
        {
```

```cpp
            assert(_regionAbove);
            ++_numSoftColumns;
        }
        else
            --hardCount;

        if (_regionAbove)
        {
            assert(_numColumns < MAX_WINDOW_WIDTH);
            if (++_numColumns == MAX_WINDOW_WIDTH)
                printf("+");  // Indicate event window overflow.
        }
    }

    // Words emitted by this region's event window are received as
    // input events by the event window of the region above, which in
    // turn sends predictions (feedback) down to this region.
    if (_regionAbove)
    {
        int oldOffset = _outputOffset;

        assert(_headColumn && _numColumns);
        if (_enablePredictions)
        {
            InitLocalPredictions();
            assert(_outputOffset);
            BuildFeedbackFilter(&_feedbackFilter);
            if (_retractedName)
            {
                _feedbackFilter.ClearBit(_retractedName);
                _retractedName = 0;
            }
            _regionAbove->ApplyFeedbackFilter(&_feedbackFilter);
        }
        SetSecondLevelScores();
        EmitWords();
        if (_enablePredictions && oldOffset > _outputOffset)
        {
            if (_outputOffset)
            {
                // Send revised feedback filter to region above.
                BuildFeedbackFilter(&_feedbackFilter);
                _regionAbove->ApplyFeedbackFilter(&_feedbackFilter);
                if (!_predictionFromAbove._seq && _outputOffset < _numColumns)
                {
                    // The region above followed the garden path to a highly unlikely prediction.
                    // Retract the last event so the region above can back out and try again.
                    UINT8 name = _regionAbove->RetractLastEvent();

                    assert(name);
                    col = _headColumn - _outputOffset;
                    if (col < _column)
                        col += ARRAY_LENGTH(_column);

                    len = col->_bestLength;
                    assert(name == col->_cell[len-1]._seq->_sequenceName);
                    _outputOffset += len;
                    BuildFeedbackFilter(&_feedbackFilter);
                    _feedbackFilter.ClearBit(name);  // Zap retracted name.
                    _regionAbove->ApplyFeedbackFilter(&_feedbackFilter);
                    if (!EmitWords())
                        _retractedName = name;
                }
            }
            else
                _regionAbove->ApplyFeedbackFilter(NULL);
        }
    }
}

//
// Called by the region below to retract the last output event that it
// sent to the region above. This function is used to back out of
// shallow garden path errors caused by bad predictions from above.
//
UINT8 MidRegion::RetractLastEvent()
{
    UINT8 event = 0;

#if SHOW_DIAGNOSTICS != 0
    printf(" (RETRACT LAST EVENT)");
#endif
    assert(_headColumn && _numSoftColumns);
    --_numSoftColumns;
    event = _headColumn->_event;
    _headColumn = !(--_numColumns) ? NULL : _headColumn = _headColumn->_prevCol;
    InitLocalPredictions();
    return event;
}

//
// Delete all junk sequences in sequence memory.
//
void MidRegion::PruneSequenceMemory()
{
```

```cpp
        for (int ix = 0; ix < ARRAY_LENGTH(_hashTable1); ++ix)
        {
            SEQUENCE **seq = &_hashTable1[ix];

            while (*seq)
            {
                if (!(*seq)->_outCount)
                {
                    SEQUENCE *temp = *seq;

                    *seq = temp->_link1;
                    delete temp;
                }
                else
                    seq = &(*seq)->_link1;
            }
        }
    }

    //
    // Convert names in input string to strings of events and print them.
    //
    void MidRegion::WriteString(UINT8 *string)
    {
        for (UINT8* name = string; *name; ++name)
        {
            UINT8 buffer[MAX_WINDOW_WIDTH + 1];
            UINT8 *str = &buffer[ARRAY_LENGTH(buffer)];
            SEQUENCE *seq;

            *--str = 0;
            for (seq = _namedSequence[*name]; seq; seq = seq->_prevSeq)
                *--str = seq->_info._event;

            assert(_regionBelow);
            _regionBelow->WriteString(str);
            printf("%s", _sequenceDelimiter);
        }
    }

    // Comparison function to use with qsort library function.
    int CompareCounts(const void *key1, const void *key2)
    {
        SEQUENCE *p = *(SEQUENCE**)key1;
        SEQUENCE *q = *(SEQUENCE**)key2;

        return (q->_outCount - p->_outCount);
    }

    //
    // For each valid sequence, build a list of probable successors.
    // Before calling this routine, call PruneSequenceMemory.
    //
    void MidRegion::BuildSuccessorLists()
    {
        int ix, jx, kx, max = 0;
        SEQUENCE *sorter[256];

        // Cache pointer to most likely sequence of length == 1.
        _rootSequence[0]._link2 = NULL;
        for (ix = 1; ix < ARRAY_LENGTH(_rootSequence); ++ix)
        {
            if (max < _rootSequence[ix]._outCount)
            {
                max = _rootSequence[ix]._outCount;
                _rootSequence[0]._link2 = &_rootSequence[ix];
            }
        }
        assert(max && _rootSequence[0]._link2);

        // After pruning, _hashTable1 contains all valid sequences of length > 1.
        memset(_hashTable2, 0, sizeof(_hashTable2));
        for (ix = 0; ix < ARRAY_LENGTH(_hashTable1); ++ix)
        {
            // In _hashTable2, create lists of valid successors for fast lookup.
            for (SEQUENCE **seq = &_hashTable1[ix]; *seq; seq = &(*seq)->_link1)
            {
                jx = GetHashIndex((*seq)->_prevSeq, 0) % ARRAY_LENGTH(_hashTable2);
                (*seq)->_link2 = _hashTable2[jx];
                _hashTable2[jx] = *seq;
            }
        }

        // In _hashTable2, sort successor lists in order of descending _outCount.
        for (ix = 0; ix < ARRAY_LENGTH(_hashTable2); ++ix)
        {
            SEQUENCE *seq = _hashTable2[ix];

            if (seq && seq->_link2)
            {
                memset(sorter, 0, sizeof(sorter));
                for (jx = 0; seq && jx < ARRAY_LENGTH(sorter); ++jx, seq = seq->_link2)
                    sorter[jx] = seq;

                assert(jx != ARRAY_LENGTH(sorter));   // Detect bad hash algorithm.
```

```cpp
            qsort(sorter, jx, sizeof(SEQUENCE*), CompareCounts);
            _hashTable2[ix] = sorter[0];
            for (kx = 0; kx < jx-1; ++kx)
                sorter[kx]->_link2 = sorter[kx+1];

            sorter[jx-1]->_link2 = NULL;
        }
    }
}

//
// Initialize predictions for all paths ending on right edge of local event window.
//
void MidRegion::InitLocalPredictions()
{
    CELL *outCell = &_localPredictions._cell[0];
    float Psow;

    memset(&_localPredictions, 0, sizeof(_localPredictions));
    outCell[0]._seq = _rootSequence[0]._link2;
    assert(outCell[0]._seq);
    Psow = (float)outCell[0]._seq->_outCount / _eventCount;
    if (!_headColumn || !_outputOffset)
    {
        // Simple case: The next character is the start of a new word.
        _outputOffset = 0;
        outCell[0]._score = Psow;
        _localPredictions._bestScore = Psow;
        _localPredictions._numCells = 1;
        _localPredictions._bestLength = 1;
        _localPredictions._event = outCell[0]._seq->info._event;
    }
    else
    {
        CELL *inCell = &_headColumn->_cell[0];
        int ix;

        // Populate the local prediction column with likely successors.
        for (ix = 0; ix < _headColumn->_numCells; ++ix)
        {
            SEQUENCE *inSeq = inCell[ix]._seq;
            int jx = GetHashIndex(inSeq, 0) % ARRAY_LENGTH(_hashTable2);
            SEQUENCE *outSeq = _hashTable2[jx];

            while (outSeq && outSeq->_prevSeq != inSeq)
                outSeq = outSeq->_link2;

            if (!outSeq)
                break;

            outCell[ix+1]._seq = outSeq;
        }
        _localPredictions._numCells = ix + 1;

        // Calculate the score for each successor in prediction column.
        for (ix = 0; ix < _headColumn->_numCells; ++ix)
        {
            SEQUENCE *inSeq = inCell[ix]._seq;
            float score = inCell[ix]._score;
            float eowCount = inSeq->_outCount - inSeq->_succCount;
            float Peow = eowCount / inSeq->_outCount;
            float Pnew = Peow * Psow;

            if (ix + 1 < _localPredictions._numCells)
            {
                SEQUENCE *outSeq = outCell[ix+1]._seq;
                float Psame = (float)outSeq->_outCount / inSeq->_outCount - Pnew;

                outCell[ix+1]._score = score * Psame;
            }
            score *= Pnew;
            if (!ix || outCell[0]._score < score)
                outCell[0]._score = score;
        }

        // Find the most likely successor in the prediction column.
        for (int ix = 0; ix < _localPredictions._numCells; ++ix)
        {
            if (!ix || _localPredictions._bestScore < outCell[ix]._score)
            {
                _localPredictions._bestScore = outCell[ix]._score;
                _localPredictions._bestLength = ix + 1;
                _localPredictions._event = outCell[ix]._seq->info._event;
            }
        }
    }
}

//
// Initialize bitmasks for use by BuildFeedbackFilter function to build
// filters.  Each bitmask indicates all the named sequences whose 1st,
// 2nd, 3rd, or 4th  event is a particular value. For a named sequence of
// length > 4, use a less storage-intensive technique to build a filter.
//
void MidRegion::InitFeedbackFiltering()
```

```cpp
    {
        assert(!_namedSequence[0]);  // Reserve 0 as the null sequence name.
        for (int name = 1; name <= _nameIndex; ++name)
        {
            SEQUENCE *seq = _namedSequence[name];
            Mask255 *mask = NULL;

            assert(name == seq->_sequenceName);
            for (int len = seq->_info._length; len; --len, seq = seq->_prevSeq)
            {
                if (len > 4)   // BUGBUG - Use named constant?
                    continue;

                mask = _eventSet[len-1][seq->_info._event];
                if (!mask)
                {
                    mask = new Mask255;
                    assert(mask);
                    _eventSet[len-1][seq->_info._event] = mask;
                }

                // Add this named sequence to the set of all named sequences
                // whose (len)th event is seq->_info._event, where len=1,2,3,4.
                mask->SetBit(name);
            }
            assert(!seq);
        }
    }

    //
    // Return true if input string starting at specified offset from
    // column col is a possible segmentation boundary. Otherwise,
    // return false. This function calls itself recursively.
    //
    bool MidRegion::IsSegmentable(int offset, COLUMN *col)
    {
        int numCells = col->_numCells;
        CELL *cell = &col->_cell[0];

        assert(offset);
        for (int len = min(offset, numCells); len; --len)
        {
            if (cell[len-1]._seq->_sequenceName ||
                col == _headColumn && cell[len-1]._seq->_precursorCount)
            {
                COLUMN *temp;

                if (offset == len)
                    return true;

                temp = col - len;
                if (temp < _column)
                    temp += ARRAY_LENGTH(_column);

                if (IsSegmentable(offset - len, temp))  // recursion
                    return true;
            }
        }
        return false;
    }

    //
    // One or more new events have entered the event window since the last
    // set of words was emitted to the region above. Build a 255-bit bitmask
    // in which bit N is set if and only if named sequence N is consistent
    // with the new events. The region above will use this bitmask to avoid
    // making predictions that are inconsistent with these new events.
    //
    bool MidRegion::BuildFeedbackFilter(Mask255 *mask)
    {
        SEQUENCE *precursor;

        assert(_outputOffset);
        if (_headColumn->_numCells < _outputOffset)
        {
            COLUMN *col = _headColumn->_prevCol;
            int len = _outputOffset;

            mask->ClearMask();
            while (--len && len > col->_numCells)
                col = col->_prevCol;

            precursor = col->_cell[len-1]._seq;
            if (precursor->_sequenceName)
                mask->SetBit(precursor->_sequenceName);
        }
        else
        {
            precursor = _headColumn->_cell[_outputOffset-1]._seq;
            assert(precursor);
            if (!precursor->_precursorCount)
            {
                // No named word starts at _outputOffset. End of stream?
                mask->ClearMask();
                return false;
```

```cpp
        }
        if (_outputOffset <= 4)  // BUGBUG - Use named constant?
        {
            // For precursor sequence of length <= 4, isolate set of
            // candidate named sequences by using bitmask operations
            // to find intersection of previously calculated supersets.
            COLUMN *col = _headColumn - _outputOffset + 1;

            if (col < _column)
                col += ARRAY_LENGTH(_column);

            assert(_eventSet[0][col->_event]);
            *mask = *_eventSet[0][col->_event];
            for (int ix = 1; ix < _outputOffset; ++ix)
            {
                col = col->_nextCol;
                assert(_eventSet[ix][col->_event]);
                mask->AndMask(_eventSet[ix][col->_event]);
            }
        }
        else
        {
            // For precursor sequence of length > 4, identify set of
            // candidate named sequences by traversing subtree in
            // memory hierarchy whose root is the precursor sequence.
            SEQUENCE *stack[MAX_WINDOW_WIDTH];
            int ix = 0;

            mask->ClearMask();
            memset(stack, 0, ARRAY_LENGTH(stack));
            stack[0] = precursor;
            do
            {
                assert(stack[ix]);
                if (stack[ix]->_sequenceName)
                    mask->SetBit(stack[ix]->_sequenceName);

                if (stack[ix]->_succCount)
                {
                    int jx = GetHashIndex(stack[ix], 0) % ARRAY_LENGTH(_hashTable2);

                    stack[++ix] = _hashTable2[jx];  // Go one level deeper.
                }
                else
                    stack[ix] = stack[ix]->_link2;

                while (ix)
                {
                    while (stack[ix] && stack[ix]->_prevSeq != stack[ix-1])
                        stack[ix] = stack[ix]->_link2;

                    // If this level is exhausted, reduce depth by one level.
                    if (stack[ix] || !--ix)
                        break;

                    assert(stack[ix]);
                    stack[ix] = stack[ix]->_link2;
                }
                assert(!ix || stack[ix]->_prevSeq == stack[ix-1]);

            } while (ix);
        }
    }

    // Include any named subsequences that are consistent with new events.
    for (SEQUENCE *seq = precursor->_prevSeq; seq; seq = seq->_prevSeq)
    {
        if (seq->_sequenceName)
        {
            int len = seq->_info._length;

            assert(len <= _outputOffset);
            if (len == _outputOffset || IsSegmentable(_outputOffset - len, _headColumn))
            {
                assert(!mask->GetBit(seq->_sequenceName));
                mask->SetBit(seq->_sequenceName);
            }
        }
    }
#if SHOW_DIAGNOSTICS != 0
    if (mask == &_feedbackFilter)
    {
        // DEBUG SPEW - Print new events used to build feedback filter.
        {
            char buffer[32];
            char *str = &buffer[ARRAY_LENGTH(buffer)];
            COLUMN *col = _headColumn;

            *--str = 0;
            for (int ix = 0; ix < _outputOffset; ++ix, col = col->_prevCol)
                *--str = col->_event;

            printf("\n%s: ", str);
        }
```

```cpp
            // DEBUG SPEW - Print out names in feedback filter.
            {
                int name = 0;
                UINT8 buf[2];

                buf[1] = '\0';
                while (name = mask->GetNext(name))
                {
                    buf[0] = name;
                    WriteString(buf);
                }
            }
        }
#endif
    return (!mask->IsEmpty());
}

//
// Called by region below to tell this region which predictions are viable.
// Viable predictions are specified as a feedback filter bitmask of up to
// 255 bits. Predictions are viable only if they are consistent with events
// the region below has already observed (bits set to 1 in the bitmask).
//
void MidRegion::ApplyFeedbackFilter(Mask255 *mask)
{
    if (mask && mask->IsEmpty())
    {
        // The region below knows of no words that contain the
        // most recent event that it received. End of stream?
        memset(&_localPredictions, 0, sizeof(_localPredictions));
        _localPredictions._bestLength = 1;
        _regionBelow->SetFeedback(0, 0.0);  // withdraw bad prediction
        return;
    }
    if (!_localPredictions._bestLength)
        InitLocalPredictions();

    if (mask)
    {
        float Psow = 0.0;
        int denom = 0;
        int max = 0;
        CELL *outCell = _localPredictions._cell;

        outCell[0]._seq = NULL;
        for (int evt = mask->GetNext(0); evt; evt = mask->GetNext(evt))
        {
            denom += _rootSequence[evt]._outCount;
            if (max < _rootSequence[evt]._outCount)
            {
                max = _rootSequence[evt]._outCount;
                outCell[0]._seq = &_rootSequence[evt];
            }
        }
        if (denom)
            Psow = (float)outCell[0]._seq->_outCount / denom;

        _localPredictions._bestScore = 0.0;
        _localPredictions._bestLength = 1;
        _localPredictions._event = 0;
        if (_headColumn)
        {
            CELL *inCell = _headColumn->_cell;

            for (int ix = 0; ix < _headColumn->_numCells; ++ix)
            {
                SEQUENCE *inSeq = inCell[ix]._seq;
                float score = inCell[ix]._score;
                float eowCount = inSeq->_outCount - inSeq->_succCount;
                float Peow = Peow = eowCount / inSeq->_outCount;
                float Pnew = Peow * Psow;

                if (ix + 1 < _localPredictions._numCells)
                {
                    float Psame;
                    SEQUENCE *outSeq;

                    // Find first viable prediction in this bucket list.
                    for (outSeq = outCell[ix+1]._seq; outSeq; outSeq = outSeq->_link2)
                        if (outSeq->_prevSeq == inCell[ix]._seq && mask->GetBit(outSeq->_info._event))
                            break;  // Found it.

                    outCell[ix+1]._seq = outSeq;
                    if (outSeq)
                    {
                        // Get sum of counts to use as denominator in probability calculation below.
                        denom = outSeq->_inCount;
                        for (outSeq = outSeq->_link2; outSeq; outSeq = outSeq->_link2)
                        {
                            if (outSeq->_prevSeq == inCell[ix]._seq && mask->GetBit(outSeq->_info._event))
                                denom += outSeq->_inCount;
                        }
                        Psame = (float)outCell[ix+1]._seq->_inCount / denom;
                        outCell[ix+1]._score = score * (1.0 - Peow) * Psame;
                        if (_localPredictions._bestScore < outCell[ix+1]._score)
```

```cpp
                    {
                        _localPredictions._bestScore = outCell[ix+1]._score;
                        _localPredictions._bestLength = ix + 2;
                        _localPredictions._event = outCell[ix+1]._seq->info._event;
                    }
                }
                else
                    outCell[ix+1]._score = 0.0;  // No viable prediction in this bucket list.
            }
            score *= Pnew;
            if (!ix || outCell[0]._score < score)
            {
                outCell[0]._score = score;
            }
        }
    }
    else
        outCell[0]._score = Psow;

    if (_localPredictions._bestScore < outCell[0]._score)
    {
        _localPredictions._bestScore = outCell[0]._score;
        _localPredictions._bestLength = 1;
        _localPredictions._event = outCell[0]._seq->info._event;
    }
#if SHOW_DIAGNOSTICS != 0
    // DEBUG SPEW - Print out prediction from upper region.
    printf("=> ");
    if (_localPredictions._event)
    {
        UINT8 buf[2];

        buf[1] = '\0';
        buf[0] = _localPredictions._event;
        _regionBelow->WriteString(buf);
    }
    else
        printf("-?- ");  // No prediction is possible.

    printf("%f ", _localPredictions._bestScore);
    if (_localPredictions._bestScore < 0.1)
        printf("  <-- NOTE LOW PROBABILITY");  // Low confidence.
#endif
    }

    // Send filtered feedback to region below.
    if (_predictionFromAbove._seq &&
        _predictionFromAbove._score > _localPredictions._bestScore &&
        _predictionFromAbove._seq->_info._length > _outputOffset)
    {
        SEQUENCE *seq = _predictionFromAbove._seq;

        while (seq->_info._length > _outputOffset + 1)
            seq = seq->_prevSeq;

        _regionBelow->SetFeedback(seq->_info._event, _predictionFromAbove._score);
    }
    else
        _regionBelow->SetFeedback(_localPredictions._event, _localPredictions._bestScore);
}

//
// Called by region above to send its prediction to this region.
//
void MidRegion::SetFeedback(UINT8 name, float score)
{
    if (!name || score < HIGHLY_UNLIKELY)
    {
        // Region above has exhausted its predictions.
        _predictionFromAbove._seq = NULL;
        _predictionFromAbove._score = 0.0;
        return;
    }
    assert(name <= _nameIndex);
    _predictionFromAbove._seq = _namedSequence[name];
    _predictionFromAbove._score = score;
}

//
// Diagnostic information: Print contents of sequence memory.
//
void MidRegion::DumpSequenceMemory()
{
    char buffer[MAX_WINDOW_WIDTH];

    memset(buffer, 0, sizeof(buffer));
    printf("\nDUMP SEQUENCE MEMORY:\n");
    for (int ix = 1; ix < ARRAY_LENGTH(_rootSequence); ++ix)
    {
        SEQUENCE *stack[MAX_WINDOW_WIDTH];
        int jx = 0;

        if (!_rootSequence[ix]._outCount)
            continue;
```

```c
            memset(stack, 0, ARRAY_LENGTH(stack));
            stack[0] = &_rootSequence[ix];
            do
            {
                assert(stack[jx]);
                buffer[jx] = stack[jx]->_info._event;
                if (stack[jx]->_succCount)
                {
                    int kx = GetHashIndex(stack[jx], 0) % ARRAY_LENGTH(_hashTable2);

                    stack[++jx] = _hashTable2[kx];   // Go one level deeper.
                }
                else
                {
                    buffer[jx+1] = '\0';
                    printf("%s\n", buffer);
                    memset(buffer, '.', jx);
                    stack[jx] = stack[jx]->_link2;
                }
                while (jx)
                {
                    while (stack[jx] && stack[jx]->_prevSeq != stack[jx-1])
                        stack[jx] = stack[jx]->_link2;

                    // If this level is exhausted, reduce depth by one level.
                    if (stack[jx] || !--jx)
                        break;

                    assert(stack[jx]);
                    stack[jx] = stack[jx]->_link2;
                }
                assert(!jx || stack[jx]->_prevSeq == stack[jx-1]);

            } while (jx);
            printf("\n");
    }
}

//
// Diagnostic information: Print all named sequences in this region.
//
void MidRegion::DumpNamedSequences()
{
    printf("\n\nNAME      SEQUENCE   SCORE\n");
    for (int ix = 1; ix <= _nameIndex; ++ix)
    {
        SEQUENCE *seq = _namedSequence[ix];
        UINT8 buffer[MAX_WINDOW_WIDTH];
        UINT8 *str = &buffer[ARRAY_LENGTH(buffer)];
        float score = seq->_accumScores / seq->_inCount;

        *--str = '\0';
        for (SEQUENCE *seq2 = seq; seq2; seq2 = seq2->_prevSeq)
            *--str = seq2->_info._event;

        printf("%3d %14s  %1.3f\n", ix, str, score);
    }
    printf("\n");
}

//
// Main program: Send stream of test data to two-level hierarchy
// of event window segmentation (EWS) algorithm instances.
//
int main()
{
    int ix, charCount = 0;
    char *s, buf[2];
    BottomRegion bottom;
    MidRegion region1(" ");   // Lower level in hierarchy
    MidRegion region2("\n");  // Higher level in hierarchy
    TopRegion top;
    Trainer train;

    // Learning phase for region1
    region1.ConnectToRegionBelow(&bottom);
    buf[1] = '\0';
    srand(123456);
    for (ix = 0; ix < NUM_LEARNING_WORDS; ++ix)
    {
        int index = rand() % ARRAY_LENGTH(_test1);

        for (s = _test1[index]; *s; ++s)
        {
            buf[0] = *s;
            region1.ProcessEvents((UINT8*)buf, 1);
        }
    }
    region1.PruneSequenceMemory();
    region1.BuildSuccessorLists();
    region1._headColumn = NULL;     // Reset event window.
#if 1
    // Use Trainer to speed up learning phase for region2.
    train.ConnectToRegionBelow(®ion1);
    for (ix = 0; ix < ARRAY_LENGTH(_test2); ++ix)
```

```
        {
            for (s = _test2[ix]; *s; ++s)
            {
                buf[0] = *s;
                region1.ProcessEvents((UINT8*)buf, 1);
            }

            // Mark the end of this training pattern.
            buf[0] = END_OF_STREAM;
            region1.ProcessEvents((UINT8*)buf, 1);
        }
        assert(region1._headColumn == NULL);
        region2.ConnectToRegionBelow(&train);   // Run training loop
        region2.ConnectToRegionBelow(®ion1);
#else
        // Propagate learning patterns for region2 through region1.
        region2.ConnectToRegionBelow(®ion1);

        // For easier debugging, make sequence names predictable.
        for (ix = 0; ix < ARRAY_LENGTH(_test2); ++ix)
        {
            for (s = _test2[ix]; *s; ++s)
            {
                buf[0] = *s;
                region1.ProcessEvents((UINT8*)buf, 1);
            }
        }

        // BUGBUG - Same NUM_LEARNING_WORDS for all regions?
        for (ix = 0; ix < NUM_LEARNING_WORDS; ++ix)
        {
            int index = rand() % ARRAY_LENGTH(_test2);

            for (s = _test2[index]; *s; ++s)
            {
                buf[0] = *s;
                region1.ProcessEvents((UINT8*)buf, 1);
            }
        }
#endif
        // Output phase
        region2.PruneSequenceMemory();
        region2.BuildSuccessorLists();
        region2._headColumn = NULL;     // Reset event window.
        region1._headColumn = NULL;
        region1._enablePredictions = true;
        region1.InitFeedbackFiltering();
        top.ConnectToRegionBelow(®ion2);
        for (ix = 0; ix < ARRAY_LENGTH(_test2); ++ix)
        {
            for (s = _test2[ix]; *s; ++s)
            {
#if SHOW_DIAGNOSTICS == 0
                UINT8 pred = bottom.GetFeedback();

                // Use upper-case letter to highlight prediction error.
                printf("%c", (pred == *s) ? *s : *s + 'A' - 'a');
#endif
                buf[0] = *s;
                region1.ProcessEvents((UINT8*)buf, 1);
                ++charCount;
            }
        }

        // Flush remaining words in event window to output stream.
        buf[0] = END_OF_STREAM;
        region1.ProcessEvents((UINT8*)buf, 1);

        // Print miscellaneous diagnostic information.
        region1.DumpNamedSequences();
#if SHOW_DIAGNOSTICS != 0
        region1.DumpSequenceMemory();
#endif
        printf("SEQUENCES ALLOCATED = %d\n", region1._allocCount);
        printf("SEQUENCES FIRING = %d\n", region1._fireCount);
        printf("TOTAL EVENT COUNT = %d\n", region1._eventCount);
        printf("SAMPLE LENGTH = %d words, %d chars\n",
                ARRAY_LENGTH(_test1), charCount);
        return 0;
}
```

The GetHashIndex function in the preceding source code listing uses the ELF hash algorithm, as described by Binstock [1].

The MidRegion class is an implementation of the EWS algorithm. The test program creates two instances of this class to serve as two consecutive levels—level 1 and level 2—in the model hierarchy. The

TopRegion and BottomRegion classes contain test instrumentation, and instances of these classes are attached to the top and bottom, respectively, of the two-level hierarchy.

The Trainer class is provided to speed up the training of level 2 so that experimentation with code changes and parameter tuning is more interactive. After level 1 finishes its training, an instance of the Trainer class captures a complete set of training patterns from level 1, and then repeatedly plays these patterns back to level 2 in random order. After level 2 finishes its training, the Trainer instance is removed so that level 1 and level 2 communicate directly with each other.

The test program can easily be configured to remove the Trainer, in which case the training patterns for level 2 all propagate through level 1. However, the increased processing in level 1 significantly increases the duration of a test run. For the simple test case used in this program, the learning accomplished by level 2 should be nearly the same regardless of whether the Trainer is used. If the number of levels in the hierarchy under test were to increase from two to three, the benefits of using the Trainer would be even more apparent.

The test program prints text to the console output device. By default, the test program is configured to indicate how successful the model is in predicting the next character in the input stream to the lower level. As previously explained in the "Discussion" section, prediction errors are highlighted as upper-case characters, and correct predictions are shown as lower-case letters. The resulting output stream looks like this:

```
FoUrscoreandsevenyearsagoOUrfathersbroughtforthonthiscontinentANEwNationCOnCe
ivedinlibertyANddedicatedtothepropositionThAtAllmenAReCreatedequal...
```

Next, the test program dumps all the character patterns in the sequence memory that the EWS algorithm in level 1 has identified as probable words. The following example shows that the character pattern "forth" is assigned the name (integer value) 11:

```
11         forth   0.833
```

The value 0.833 on the right is the probability that an instance of the sequence f-o-r-t-h in the input stream to level 1 is, in fact, an occurrence of the word "forth"—for example, it might instead be an instance of the word "for" followed by a word that starts with the letters "th".

The test program can be configured to print a different set of results by changing the value of constant SHOW_DIAGNOSTICS from 0 to 1. The following is an example of the resulting output:

```
ha: have hallow => hallow 0.998310
```

The letters "ha" to the left of the colon are the current contents of the event window in level 1. The two words "have" and "hallow" are the only words represented in the feedback filter that level 1 sends to level 2. The word "hallow" to the right of the arrow is the word predicted by level 2, and the number 0.998310 on the right is the probability estimated by level 2 that this prediction is correct.